\documentclass{article}
\usepackage{adjustbox}
\usepackage{tcolorbox}
\usepackage{xcolor}

\usepackage{microtype}
\usepackage{graphicx}
\usepackage{subcaption}
\usepackage{booktabs} %
\definecolor{lightblue}{rgb}{0.6, 0.8, 0.9}
\definecolor{darkblue}{rgb}{0.2,0.4,0.6}
\definecolor{lightgreen}{rgb}{0.5, 0.8, 0.6}
\definecolor{darkgreen}{rgb}{0, 0.55, 0.12}
\definecolor{darkred}{rgb}{0.6,0,0}
\usepackage{amsmath,amsfonts,bm}

\def\eqref#1{equation~\ref{#1}}
\def\1{\bm{1}}

\DeclareMathAlphabet{\mathsfit}{\encodingdefault}{\sfdefault}{m}{sl}
\SetMathAlphabet{\mathsfit}{bold}{\encodingdefault}{\sfdefault}{bx}{n}

\usepackage{algorithmic}
\usepackage{algorithm}
\usepackage{arydshln}

\usepackage{hyperref}
\let\OriginalAddContentsLine\addcontentsline

\usepackage[accepted]{icml2026}

\usepackage{amsmath}
\usepackage{amssymb}
\usepackage{mathtools}
\usepackage{amsthm}
\usepackage{enumitem}
\usepackage{bm}
\usepackage{url}

\usepackage{pifont}
\usepackage{booktabs}
\usepackage{multicol}
\usepackage{multirow}
\usepackage{listings}
\usepackage{balance}
\usepackage{needspace}
\usepackage{xspace}
\usepackage{colortbl}
\usepackage{setspace}
\usepackage{fancybox}
\usepackage{tocloft}
\usepackage{etoc}

\usepackage[capitalize, noabbrev]{cleveref}

\theoremstyle{plain}

\allowdisplaybreaks

\newtheorem{thm}{Theorem}

\newtheorem{prop}[thm]{Proposition}
\newtheorem{cor}[thm]{Corollary}

\theoremstyle{definition}

\theoremstyle{definition}

\theoremstyle{remark}

\usepackage[textsize=tiny]{todonotes}

\icmltitlerunning{Calibrated Multimodal Representation Learning with Missing Modalities}

\begin{document}
  \twocolumn[ \icmltitle{Calibrated Multimodal Representation Learning with Missing Modalities}

  \icmlsetsymbol{equal}{*}

  \begin{icmlauthorlist}
    \icmlauthor{Xiaohao Liu}{nus} \icmlauthor{Xiaobo Xia}{ustc} \icmlauthor{Jiaheng Wei}{hkust} \icmlauthor{Shuo Yang}{hits} \icmlauthor{Xiu Su}{csu} \icmlauthor{See-Kiong Ng}{nus} \icmlauthor{Tat-Seng Chua}{nus}
  \end{icmlauthorlist}
  
  \icmlaffiliation{nus}{National University of Singapore} 
  \icmlaffiliation{ustc}{University of Science and Technology of China} \icmlaffiliation{hkust}{The Hong Kong University of Science and Technology (Guangzhou)}
  \icmlaffiliation{hits}{Harbin Institute of Technology (Shenzhen)}
  \icmlaffiliation{csu}{Central South University}

  \icmlcorrespondingauthor{Xiaobo Xia}{xiaoboxia@ustc.edu.cn}

  \icmlkeywords{Machine Learning, ICML}
  
  \vskip 0.3in ]

  \printAffiliationsAndNotice{} 

\begin{abstract}
    Multimodal representation learning harmonizes distinct modalities by aligning them into a unified latent space. Recent research generalizes traditional cross-modal alignment to produce enhanced multimodal synergy but requires all modalities to be present for a common instance, making it challenging to utilize prevalent datasets with missing modalities. We provide theoretical insights into this issue from an \textit{anchor shift} perspective. Observed modalities are aligned with a local anchor that deviates from the optimal one when all modalities are present, resulting in an inevitable shift.
    To address this, we propose CalMRL to calibrate incomplete alignments caused by missing modalities. CalMRL leverages the priors and the inherent connections among modalities to model the imputation for the missing ones at the representation level. To resolve the optimization dilemma, we employ a bi-step learning method with the closed-form solution of the posterior distribution of shared latents. We validate its mitigation of anchor shift and convergence with theoretical guidance. By equipping the calibrated alignment with the existing advanced method, we offer new flexibility to absorb data with missing modalities, which is originally unattainable. Extensive experiments demonstrate the superiority of CalMRL. The code is released at \url{https://github.com/Xiaohao-Liu/CalMRL}.
\end{abstract}

% Lay of summary: The real world is a mix of different modalities that work together to give us a complete picture. Models try to align these modalities into a shared understanding. However, real-world data is often messy and incomplete, with missing modalities. This creates a problem we call "anchor shift". We develop a calibration method called CalMRL. By modeling how different modalities relate to one another, CalMRL imputes the missing ones, corrects the bias, and keeps the alignment accurate. Our work explains why this drift happens and provides a practical way to fix it, allowing AI to learn more effectively from the imperfect, multi-sensory data found in the real world.

  \section{Introduction}

\begin{figure}[t]
    \centering
    \includegraphics[width=\linewidth]{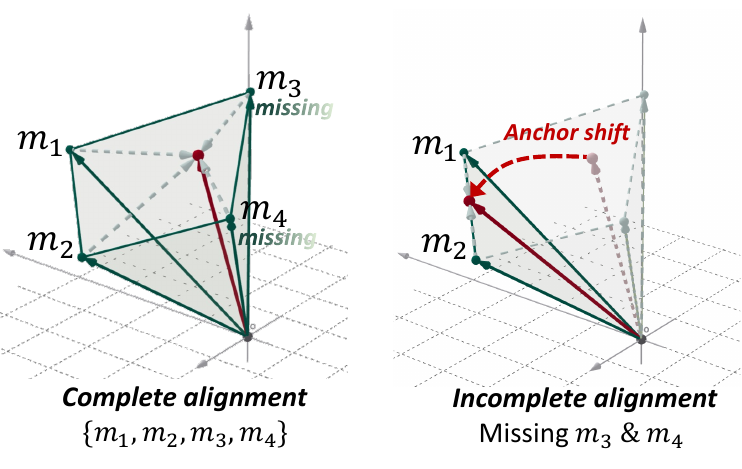}
    \vspace{-6mm}
    \caption{\textbf{Missing modalities result in distorted representation alignment.} Different modalities (in green) are aligned together with a virtual anchor (in red) implicitly with all modalities present. With missing modalities, observed ones are enforced to be aligned with a local anchor, deviating from the correct, \textit{i.e.}, anchor shift. }
    \label{fig:top}
    \vspace{-5mm}
\end{figure}

The empirical world is perceived by humans through diverse modalities, \textit{e.g.}, vision, audio, and text~\cite{ngiam2011multimodal,lu2023theory, xu2023multimodal, cao2025generalized,liang2021multibench,xin2025lumina,lei2024vit,gao2026omnimodal,lv2026spd}. For example, eyes capture the shape or colors of one real instance, ears detect, and human language records. Different reflections on the instance shape its abstract form \cite{huh2024position, tjandrasuwita2025understanding}, which connects heterogeneous modalities. The single sound you heard invokes your rough imagination in the brain. 
Multimodal representation learning \cite{radford2021learning, girdhar2023imagebind, chen2023vast, cicchetti2024gramian, dufumier2024align, liu2026principled}, a key topic in multimodal learning, leverages aligned data to harmonize distinct unimodal encoders and bridges modalities, operating under this philosophy. 

Recent research learns multiple modalities simultaneously, rather than grounding one to another \cite{radford2021learning, wu2022wav2clip, xu2021videoclip, luo2022clip4clip,guzhov2022audioclip, zhang2022pointclip}, fostering greater synergy among them \cite{cicchetti2024gramian,giordano2025triangle,liu2026principled}. To achieve multimodal alignment, they utilize geometric techniques to pull different unimodal representations together, ideally achieving a similar point (\textit{i.e.}, anchor). 
Unfortunately, it requires training on datasets with all modalities present, \textit{i.e.}, a complete set of modalities for one instance, thus ensuring an unbiased alignment. This introduces an inevitable challenge: collecting such comprehensive datasets is costly and contrasts with the prevalence of incomplete modality data, like paired data (\textit{e.g.}, vision-text \cite{jia2009imagenet, yi2024internvid, kepan2025openvid} or audio-text \cite{kim2019audiocaps, drossos2020clotho, wu2023large}). The disparity exists between different data, hindering a collective effect. 
In this paper, we frame this as a \textit{missing modality} problem that is underexplored in multimodal representation learning. 

Prior studies, like ImageBind \cite{girdhar2023imagebind} and LangugeBind \cite{zhu2023languagebind}, integrate several paired data (\textit{i.e.}, incomplete modality data), using one existing modality (\textit{i.e.}, vision or text) as the aligning anchor \cite{guo2023point,chen2023vast, lyu2024unibind, wang2024omnibind}\footnote{We provide the full literature review in Appendix~\ref{sec:related_work}.}. To ensure the learning stability, the parameters of the corresponding encoder are fixed. They achieve aggregation of diverse modalities, yet bottlenecking the mutual improvements, and heavily depending on the performance of the fixed anchor. The issue of missing modalities is normally bypassed and set aside. 

We delve into addressing this and analyzing with a natural perspective, anchor shifting. As illustrated in Figure~\ref{fig:top}, for complete alignment when all modalities are present, unimodal representations converge toward a virtual anchor within the space spanned by all modalities. In cases of incomplete alignment (\textit{i.e.}, missing modalities $m_3$ and $m_4$), observed modalities are aligned with a local anchor, deviating from the optimal one associated with the instance, \textit{i.e.}, resulting in \textit{anchor shifting}.
Reflecting on human perception, although we cannot touch or see a specific instance, we can still roughly connect it to other modalities via compensating for the missing ones, rather than being confined to solely what we can perceive. 
The compensation capability of humans inspires us to leverage the priors to impute missing modalities, thus calibrating the virtual anchor. 

To this end, we introduce a novel multimodal representation learning framework, named \textbf{CalMRL}, which calibrates alignment in the presence of missing modalities. Specifically, we focus on the missing modality problem in the representation learning phase and theoretically analyze the inevitability of anchor shift.
The main idea comes to light: construct modalities if they are missing, thus compensating for the anchor shift. 
We leverage the priors of missing modalities and the inherent connections among modalities, and introduce a simple generative model, where modalities share common latents, and meanwhile, possess their own distinctness. 
To achieve this, we adopt a two-step learning method. We first derive a closed-form solution for the posterior distribution of the shared latents with fixed parameters. Then, using this posterior, we optimize the generative parameters. By iterating these two steps, we can progressively refine the parameters, updating only with the observed modalities. 
Missing modalities can be imputed by the shared latents, derived from the observations, and their priors. 
We further provide theoretical evidence to support the minor difference between the real and the imputed, and the convergence of our method.
This method enables learning a calibrated alignment. We concatenate the observed and imputed modal representations and jointly optimize them with the objective of PMRL~\cite{liu2026principled}, aligning all modalities simultaneously. 
To validate the efficacy and rationale of CalMRL, we conduct extensive experiments and empirical analysis by comparing with state-of-the-art~(SOTA) methods. 
Before delving into details, we summarize our contributions as follows:
\begin{itemize}
    \item We introduce Calibrated Multimodal Representation Learning (CalMRL) to address the overlooked missing modalities dilemma, which leads to anchor shift in our theoretical analysis.
    \item To calibrate the alignment, we propose using a generative model to impute missing modalities, thus compensating for anchor shift. We refine the imputation precision by iterating the posterior inference and parameter optimization with theoretical grounding, followed by optimizing encoders by incorporating both observed and imputed modalities.
    \item We conduct extensive experiments to demonstrate the superiority of CalMRL and provide strong empirical evidence supporting our design rationale. Comprehensive empirical studies and discussions are also provided.
\end{itemize}

\section{Preliminaries}

\noindent\textbf{Aligning all modalities.} Different modalities produce synergies, presenting inherent connections despite the heterogeneity. Such connections serve as the key assumption to align different modalities together to learn multimodal representations~\cite{lu2023theory, liang2023quantifying}. Contrastive learning successfully optimizes the similarities of modality pairs \cite{radford2021learning,girdhar2023imagebind}. Recent work proposes manipulating the singular value of a GRAM matrix to align multimodalities simultaneously \cite{liu2026principled}. Specifically, all the uni-modal representations are concatenated together, \textit{i.e.}, $\mathbf{z} = [\mathbf{z}^{m_1},\cdots, \mathbf{z}^{|\mathcal{M}|}]$. $\sigma_1$ denotes the largest singular values of the GRAM matrix $\mathbf{z}\mathbf{z}^\top$. By maximizing the largest one $\sigma_1$ among others $\{\sigma_i\}_{>1}$, it excels pair-wise methods in alignment tasks, yet is limited by predefined modalities, hard to extend for arbitrary modalities, and can be enhanced with modality-missing datasets.

\noindent\textbf{Modality missing.} Given the observed multimodal features $\mathbf{X}^\Omega$, the latent/unified representation can be derived via $p(\widetilde{\mathbf{z}}|\mathbf{X}^\Omega) = p(\widetilde{\mathbf{z}}| \{\mathbf{x}^m\}_{m\in\Omega})$. $\Omega$ is the set of observed modalities and $\Omega\subseteq\mathcal{M}$, where $|\mathcal{M}|=k$. If $|\Omega|<k$, it incurs the phenomena of \textit{modality missing}, which is quite common that some multimodal datasets only contain 2 modalities~\cite{jia2009imagenet, zhu2023languagebind}, while some involve more~\cite{chen2023vast}. For instance, ImageNet~\cite{jia2009imagenet} only contains vision and text, while the VAST dataset~\cite{chen2023vast} additionally includes audio and subtitles. Our formulation is designed at the lower (instance) level to cover different cases, where we can view a dataset as a set of instances that miss the same modalities. That 
$\widetilde{\mathbf{z}}$ indicates the shared information across modalities. This further informs that the independence among modalities conditioned by $\widetilde{\mathbf{z}}$, formally, $\mathbf{x}^{m}\perp\mathbf{x}^{m'} | \widetilde{\mathbf{z}}, \forall \{m,m'\}\subset\mathcal{M}$. 

\noindent\textbf{Probabilistic PCA.} It introduces a generative model which works as $\mathbf{x} = \mathbf{W}\mathbf{z} +\boldsymbol{\mu} + \boldsymbol{\epsilon}$, where $\boldsymbol{\mu}\in\mathbb{R}^{d'}$ and $\mathbf{z}\in\mathbb{R}^d$ follows $\mathcal{N}(\mathbf{0}, \mathbf{I})$~\cite{tipping1999probabilistic,ghojogh2021factor}. $\mathbf{W}\in \mathbb{R}^{d'\times d}$ transforms the latents $\mathbf{z}$ into observed space and $\boldsymbol{\epsilon}\sim \mathcal{N}(\mathbf{0}, \sigma^2\mathbf{I})$. Therefore, the conditional distribution of the observed variable is $p(\mathbf{x}|\mathbf{z}) = \mathcal{N}(\mathbf{W}\mathbf{z}+\boldsymbol{\mu}, \sigma^2\mathbf{I})$, and the marginal distribution is $p(\mathbf{x}) = \mathcal{N}(\boldsymbol{\mu}, \mathbf{W}\mathbf{W}^\top + \sigma^2\mathbf{I})$. The closed-form solution is $\boldsymbol{\mu} = \frac{1}{N}\sum_{i=1}^{M} \mathbf{x}_n$, $\sigma^2 = \frac{1}{d'-d}\sum_{i=d+1}^{d'}\lambda_i$, and $\mathbf{W} = \mathbf{U}_d(\Lambda - \sigma^2\mathbf{I})^{1/2}$. $\mathbf{U}\mathbf{\Lambda}\mathbf{U}^\top = \mathrm{SVD}(\mathbf{S})$ decomposes the data covariance $\mathbf{S}$ to eigenvectors $\mathbf{U}$ and eigenvalue matrix $\mathbf{\Lambda} = \mathrm{diag}(\lambda_1, \dots, \lambda_{d'})$. We use similar modeling principles, specializing it for multimodal scenarios and emphasizing the connections and distinctiveness of modalities. 
  \section{Calibrated Multimodal Representation Learning}

In this section, we start with the formulation of the incomplete alignment, where the instance misses some modalities (\textit{e.g.}, a text-audio pair missing its visual component), which also leads to the phenomenon of anchor shift. The absence of a modality causes the alignment center to deviate from the optimal ``complete'' multimodal space. We systematically analyze this problem and propose CalMRL by introducing an imputation mechanism that imputes representations for missing modalities in a latent space, effectively calibrating the alignment. We also provide a rigorous theoretical analysis of the alleviating effects and convergence analysis for the overall optimization objective.

\begin{figure*}
    \centering
    \vspace{3mm}
    \includegraphics[width=0.95\linewidth]{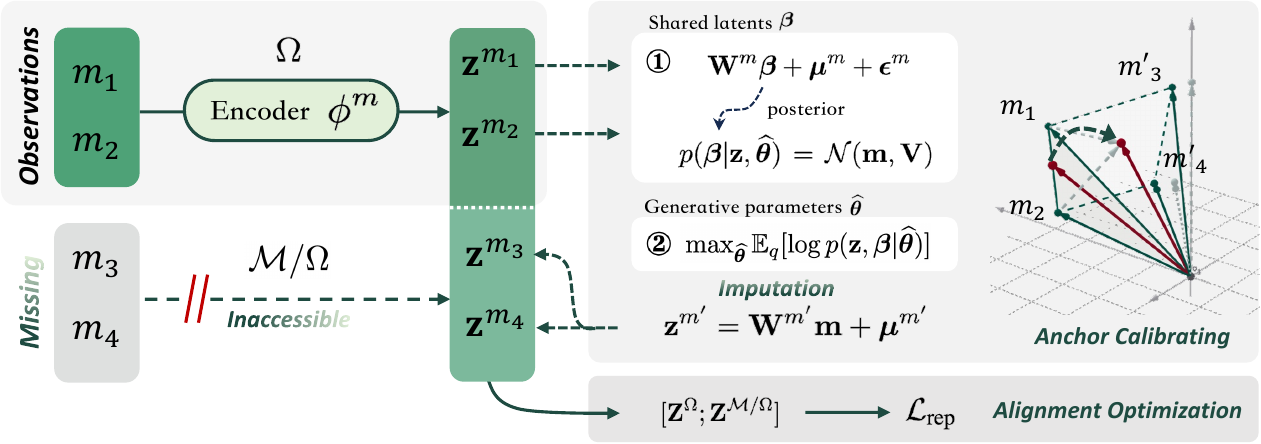}
    \caption{\textbf{The overall framework of CalMRL.} Observed unimodal content is first encoded to corresponding representations $\{\mathbf{z}^m\}_{m\in\Omega}$ with individual encoders ${\phi^m}$ in $\boldsymbol{\theta}$. Despite the missing modalities (\textit{i.e.}, $\mathcal{M}/\Omega$), CalMRL calibrates multimodal alignment whereby missing modalities are imputed by generative parameters $\widehat{\boldsymbol{\theta}}$. Finally, $\mathcal{L}_{\text{rep}}$ optimizes the observed unimodal encoder to be aligned with the calibrated direction.}
    \label{fig:framwork}
    \vspace{-8mm}
\end{figure*}

\noindent\textbf{Incomplete alignment.} We consider a practical scenario where we cannot collect datasets with complete modalities. Formally, we represent the GRAM matrix as follows:
\begin{align}\mathbf{G}^\Omega &= \begin{bmatrix}
 1& \langle \mathbf{z}^{m_1},\mathbf{z}^{m_2} \rangle & \cdots & \langle \mathbf{z}^{m_1},\mathbf{z}^{m_{k'}} \rangle \\
\langle \mathbf{z}^{m_2},\mathbf{z}^{m_1} \rangle & 1 & \cdots & \langle \mathbf{z}^{m_2},\mathbf{z}^{m_{k'}} \rangle \\
\vdots & \vdots & \ddots & \vdots \\
\langle \mathbf{z}^{m_{k'}},\mathbf{z}^{m_1} \rangle & \langle \mathbf{z}^{m_{k'}},\mathbf{z}^{m_2} \rangle & \cdots & 1
\end{bmatrix} \nonumber\\
&= \phi(\mathbf{X}^\Omega)\phi(\mathbf{X}^\Omega)^\top,
\end{align}
where $\mathbf{G}^\Omega$ is one of the block matrices belonging to the complete one $\mathbf{G}\in \mathbb{R}^{k\times k}$. In practice, when training on a mix of data, it is hard to ensure that all the data instances have the same modalities involved. In other words, one contains vision and text pairs, while another might contain audio and text data. They all miss at least one modality of data to implement the complete alignment. 

\noindent\textbf{Anchor shift $\boldsymbol{\Delta}(\mathbf{Z}, \mathbf{Z}^\Omega)$.} Multiple modalities are aligned together with a virtual centroid (\textit{i.e.}, anchor). Intuitively, there is a deviation between anchors in incomplete modalities and the complete ones. To confirm this, we also provide the theoretical evidence for the anchor shift. 
\begin{thm}[Anchor shift under incomplete modality alignment]
\label{thm:anchor_shift}
Let $\mathbf{u}_1$ and $\mathbf{u}_1^\Omega$ be the leading left singular vectors of the full multimodal matrix $\mathbf{Z}$ and its observed submatrix $\mathbf{Z}^\Omega$, respectively. Define $\sigma_1 = \|\mathbf{Z}\|_2$, $\sigma_1^\Omega = \|\mathbf{Z}^\Omega\|_2$, and  
$\eta := \sqrt{\sum_{m \in \bar{\Omega}} \langle \mathbf{u}_1^\Omega, \mathbf{z}^m \rangle^2}$.
Then the anchor shift satisfies  
\begin{equation}
    \sqrt{2\left(1 - \frac{\sigma_1^\Omega + \eta^2}{\sigma_1}\right)} \;\le\; \underbrace{\|\mathbf{u}_1 - \mathbf{u}_1^\Omega\|}_{\|\boldsymbol{\Delta}\|} \;\le\; \frac{\sqrt{2}\,\|\mathbf{Z}^{\bar{\Omega}}\|_2}{\sigma_1 - \sigma_2}. \nonumber
\end{equation}
\end{thm}
\begin{tcolorbox}[colframe=gray!40, colback=gray!10, boxrule=0.2pt, arc=1pt, boxsep=0pt, left=2pt, right=2pt, top=2pt, bottom=2pt]
\textbf{Remark.} The lower bound is \textit{strictly positive} whenever the missing modalities contribute non-zero alignment ($\eta > 0$) or the observed data fails to capture the full leading singular value ($\sigma_1^\Omega < \sigma_1$). This implies that \textit{any modality loss inevitably perturbs the virtual anchor}, no matter how well the remaining modalities are aligned. The upper bound shows that this perturbation can be mitigated, but never eliminated, by a large spectral gap (\textit{i.e., strong alignment}) and limited energy in the missing modalities.
\end{tcolorbox}

\noindent\textbf{Missing modalities via imputation.} 
We design the generation of unimodal representation to maintain intermodality connections following the conventional principles \cite{lu2023theory, gupta2025better}:
\begin{align}
    \mathbf{z}^m = \mathbf{W}^m\boldsymbol{\beta}+ \boldsymbol{\mu}^m +\boldsymbol{\epsilon}^m, \boldsymbol{\epsilon}^m\sim \mathcal{N}(\mathbf{0}, (\sigma^{m})^2\mathbf{I}).
    \label{eq:generative_model}
\end{align}
The representation for modality $m$ is conditioned by shared latents $\boldsymbol{\beta}$ and its uniqueness $\boldsymbol{\mu}_m$. In this case, we can impute the missing modalities at the representation level to compensate for the anchor offset, leading to $\boldsymbol{\Delta}(\mathbf{Z}, [\mathbf{Z}^\Omega;\mathbf{Z}^{\bar{\Omega}}]) < \boldsymbol{\Delta}(\mathbf{Z}, \mathbf{Z}^\Omega)$. To this end, even with the missing modalities, the observed modalities can be aligned in an approximated environment with complete modalities. 

To optimize the parameters, we aim to maximize the marginal log-likelihood:
\begin{align}
   \arg\max_{\widehat{\boldsymbol{\theta}}}\sum_{i=1}^N\log p(\mathbf{z}) = \arg\max_{\widehat{\boldsymbol{\theta}}}\sum_{i=1}^N\log \int p(\mathbf{z} \mid \boldsymbol{\beta}) p(\boldsymbol{\beta}) \, d\boldsymbol{\beta},
\end{align}

where $N$ denotes the batch size and $\widehat{\boldsymbol{\theta}} = \{\mathbf{W}^m,\boldsymbol{\mu}^m, \sigma^m\}_{m\in\mathcal{M}}$.
Unfortunately, as different unimodal parameters are hinged with the common $\boldsymbol{\beta}$, we cannot resolve the parameters with a closed-form solution via naive Probabilistic PCA. 
For each example, we introduce an arbitrary distribution $q(\boldsymbol{\beta})$ and derive the evidence lower bound:
\begin{align}
    \log p(\mathbf{z}) = \log\mathbb{E}_{q}\Big[\frac{p(\mathbf{z},\boldsymbol{\beta})}{q(\boldsymbol{\beta})} \Big]\ge \mathbb{E}_{q}\log\Big[\frac{p(\mathbf{z},\boldsymbol{\beta})}{q(\boldsymbol{\beta})} \Big].
    \label{eq:ELBO}
\end{align}
By iteratively optimizing $q$ and $\widehat{\boldsymbol{\theta}}$, we can elevate the likelihood. 

\noindent\textbf{Resolving $\widehat{\boldsymbol{\theta}}$.}
Accordingly, we obtain the generative parameters with a bi-step optimization.

\ding{172} Fixing the parameters, we optimize the lower bound concerning $q$,  \textit{i.e.},
    $\max_q\mathbb{E}_q[\log p(\mathbf{z}, \boldsymbol{\beta} | \widehat{\boldsymbol{\theta}})] - \mathbb{E}_q[\log q(\boldsymbol{\beta})]$. 
The equality of Eq.~(\ref{eq:ELBO}) holds if and only if $q(\boldsymbol{\beta}) = p(\boldsymbol{\beta} | \mathbf{z}, \widehat{\boldsymbol{\theta}})$. Besides, we can obtain the posterior $p(\boldsymbol{\beta} | \mathbf{z}, \widehat{\boldsymbol{\theta}}) = \mathcal{N}(\mathbf{m}, \mathbf{V})$ 
in a Gaussian distribution. In particular, 
\begin{equation}
    \begin{aligned}
        \mathbf{V} &= \Big[\mathbf{I} + \sum_{m\in\Omega} \frac{1}{(\sigma^m)^2}\mathbf{W}^{m\top}\mathbf{W}^m \Big]^{-1},\\
        \mathbf{m} &= \mathbf{V}\sum_{m\in\Omega} \frac{1}{(\sigma^m)^2}\mathbf{W}^{m\top} (\mathbf{z}^m - \boldsymbol{\mu}^m),
    \end{aligned}
    \label{eq:poterior}
\end{equation}
and we can update \textit{only} with the observed modalities in $\Omega$. 

\ding{173} Given the posterior, we can maximize the lower bound concerning $\widehat{\boldsymbol{\theta}}$, \textit{i.e.}, $\max_{\widehat{\boldsymbol{\theta}}}\mathbb{E}_q[\log p(\mathbf{z}, \boldsymbol{\beta} | \widehat{\boldsymbol{\theta}})]$. 
Here we derive the closed-form solution for each parameter as follows:

{\small
\begin{align}
    \left\{
    \begin{aligned}
        \boldsymbol{\mu}^m &= \frac{1}{N}\sum_{i=1}^N (\mathbf{z}^m_i - \mathbf{W}^m\mathbf{m}_i),\\
     \mathbf{W}^m &= \left( \sum_{i=1}^N (\mathbf{z}_i^{m} - \boldsymbol{\mu}^{m}) \mathbf{m}_i^\top \right) \left( \sum_{i=1}^N \mathbb{E}[\boldsymbol{\beta}_i \boldsymbol{\beta}_i^\top] \right)^{-1},\\
     (\sigma^m)^2 &= \frac{1}{N d} \sum_{i=1}^N \Big[ \| \mathbf{z}_i^{m} - \boldsymbol{\mu}^{m} - \mathbf{W}^{m} \mathbf{m}_i \|^2 \\
     & \quad + \mathrm{Tr}[ \mathbf{W}^{m\top} \mathbf{W}^{m} \mathbf{V}] \Big].
    \end{aligned}
    \right.
    \label{eq:hat_theta}
\end{align}
}
Alternatively, the parameters of the model can be optimized by minimizing the Gaussian negative log-likelihood loss. 
These generative parameters can be refined by the next iteration. For the generation of representation for missing modalities $\mathbf{z}^{m'}$, we have the following proposition.

\begin{prop}[Missing modality imputation]
Given observed modalities \(\Omega\) and learned generative parameters \(\widehat{\boldsymbol{\theta}} = \{\mathbf{W}^m, \boldsymbol{\mu}^m, \sigma^m\}_{m\in\mathcal{M}}\), the representation of any missing modality \(m' \in \mathcal{M}\setminus\Omega\) can be imputed in closed form as  
\begin{align}
    \widehat{\mathbf{z}}^{m'} = \mathbf{W}^{m'}\mathbf{m} + \boldsymbol{\mu}^{m'}, \quad \forall m'\in \mathcal{M}/\Omega.
\end{align}
\end{prop}
\vspace{-2mm}
We provide the detailed derivation in Appendix~\ref{sec:appendix:resolving_theta}. 

\noindent\textbf{Training objective.} For the next phase, we optimize the encoders $\phi: \mathcal{X}\to \mathcal{Z}$ with parameters $\boldsymbol{\theta}$. We complete the multimodal representations and execute the maximum singular values following \cite{liu2026principled}.
{\small\begin{align}
    \mathbf{U\Lambda}\mathbf{V} = \mathrm{SVD}([\mathbf{Z}^{\Omega}; \widehat{\mathbf{Z}}^{\mathcal{M}/\Omega}]), \quad \mathbf{\Lambda} = \mathrm{diag}(\lambda_1,\dots, \lambda_k).
\end{align}}
Accordingly, the learning objective is designed to maximize the largest singular value to enforce full alignment, and the corresponding eigenvectors are used for instance-level regularization, ensuring uniformity, as follows:
{\small
\begin{align}
    \mathcal{L}_{\text{rep}} = -&\frac{1}{N} \sum_{i=1}^N \Big[\frac{\exp[\lambda_1 /\tau]}{\sum_{j=1}^k\exp[\lambda_j /\tau]}+\frac{\exp[\mathbf{u}_1^{i\top}\mathbf{u}_1^i /\tau']}{\sum_{j=1}^N\exp[\mathbf{u}_1^{i\top}\mathbf{u}_1^j /\tau']} \Big] \nonumber\\
    & + \alpha\cdot\mathbb{E}_{\{m\}^k\sim\mathcal{M}}[y\log\hat{y} + (1-y)\log(1-\hat{y})],
    \label{eq:loss}
\end{align}
}
where $N$ denotes the number of data points in a batch. The instance matching loss is applied only to observed modalities, and weighted by $\alpha=0.1$ with a multimodal encoder and an MLP layer to predict whether the multimodal data is matched or not, returning the prediction $\hat{y}$ \cite{chen2023vast, liu2026principled,  giordano2025triangle}.

\vspace{-1mm}
\subsection{Analysis}
\vspace{-1mm}

To shed light on our method, we provide a more in-depth analysis of how it alleviates the anchor shift and whether it can converge.

\noindent\textbf{Alleviating anchor shift.} CalMRL introduces the imputation of missing modalities to mitigate the anchor shift in multimodal alignment. We derive the comparison results of the anchor offset before and after calibration as follows.
\begin{cor}[Less anchor shift with calibration]
\label{cor:less_anchor_shift}
Let each imputation satisfies $\|\widehat{\mathbf{z}}^{m'} - \mathbf{z}^{m'}\|_2 \le \varepsilon,\forall m'\in \bar{\Omega}$. Then the calibration-induced anchor deviation obeys $\|\boldsymbol{\Delta}(\mathbf{Z}, [\mathbf{Z}^\Omega;\mathbf{Z}^{\bar{\Omega}}])
\|< \|\boldsymbol{\Delta}(\mathbf{Z}, \mathbf{Z}^\Omega)\|$ if and only if 
\begin{align}
    \varepsilon < \frac{\sigma_1 - \sigma_2}{\sqrt{|\bar{\Omega}|}}
\sqrt{1 - \frac{\sigma_1^\Omega + \eta^2}{\sigma_1}}.
\end{align}
\end{cor}
\begin{tcolorbox}[colframe=gray!40, colback=gray!10, boxrule=0.2pt, arc=1pt, boxsep=0pt, left=2pt, right=2pt, top=2pt, bottom=2pt]
\textbf{Remark.} 
This corollary reveals that the benefit of calibration for anchor shift, especially in the case of strong alignment among modalities ($\uparrow \sigma_1 - \sigma_2$) and relatively small contribution of missing modalities ($\downarrow \frac{\sigma_1^\Omega + \eta^2}{\sigma_1}$). 
\end{tcolorbox}
\vspace{3mm}

\noindent\textbf{The convergence analysis.} CalMRL utilizes the bi-step iteration to optimize the generative parameters, which raises concerns about its convergence. To answer this, we confirm the monotonicity of the log-likelihood as follows.
\begin{cor}[Monotonicity for CalMRL imputation]
Let $\widehat{\boldsymbol{\theta}}^{(t)}$ be the parameters at iteration $t$ applied to the observed-data log-likelihood $L(\widehat{\boldsymbol{\theta}}) = \sum_n \log p(\mathbf{z}_n^\Omega \mid \widehat{\boldsymbol{\theta}})$ under the generative model in Eq.~(\ref{eq:generative_model}). Given the solution of $q$ and $\widehat{\boldsymbol{\theta}}$, $L(\widehat{\boldsymbol{\theta}}^{(t+1)}) \ge L(\widehat{\boldsymbol{\theta}}^{(t)})$.  
\end{cor}
\begin{tcolorbox}[colframe=gray!40, colback=gray!10, boxrule=0.2pt, arc=1pt, boxsep=0pt, left=2pt, right=2pt, top=2pt, bottom=2pt]
\textbf{Remark.} 
The monotonic increase in likelihood guarantees stable convergence of the generative model, generally to a local stationary point (see proof in Appendix~\ref{sec:appendix:convergence_analysis}).  In CalMRL, this subroutine operates within a larger alternating-optimization framework that updates the unimodal encoders, thereby providing a solid foundation for the overall convergence of the full training procedure.
\end{tcolorbox}

\begin{algorithm}[t]
\caption{CalMRL (Training)}
\label{alg:promrl}
\begin{algorithmic}[1]
\REQUIRE Multimodal dataset $\{ \mathbf{x}_i \}_{i=1}^N$; 
         encoder parameters $\boldsymbol{\theta}$;  
         generative parameters $\widehat{\boldsymbol{\theta}}$;  
         $\tau, \tau'$; $\alpha$
\ENSURE  $\phi_{\boldsymbol{\theta}}$ and $\widehat{\boldsymbol{\theta}}$

\STATE Initialize parameters $\phi_{\boldsymbol{\theta}}$ and  $\widehat{\boldsymbol{\theta}}$

\FOR{each batch $\mathcal{B} = \{ \mathbf{x}_i \}_{i=1}^N$}
    \STATE Encode: $\mathbf{z}_i^m = \phi_{\boldsymbol{\theta}}^m(\mathbf{x}_i^m),\ \forall m \in \Omega_i$
    \STATE \begin{tcolorbox}[colframe=gray!20, colback=gray!10, coltitle=black,title=\textbf{Missing modalities via imputation}, boxrule=1pt, arc=2pt, boxsep=1pt, left=2pt, right=0pt, top=-10pt, bottom=0pt]
        \begin{equation}
         \text{Update: }\left\{\begin{aligned}
            &\mathbf{m}_i, \mathbf{V}_i \gets  \text{Eq.~(\ref{eq:poterior})}\\
            &\widehat{\boldsymbol{\theta}} = \{\mathbf{W}^m, \boldsymbol{\mu}^m, \sigma^m \}_{\Omega} \gets  \text{Eq.~(\ref{eq:hat_theta})}\quad\quad\quad
        \end{aligned}
        \right.\nonumber
    \end{equation}
    Impute: $\mathbf{z}_i^{m'} \gets \mathbf{W}^{m'} \mathbf{m}_i + \boldsymbol{\mu}^{m'}, m' \in \mathcal{M} \setminus \Omega_i$
    \end{tcolorbox}
    \STATE  Update $\boldsymbol{\theta}$: $\boldsymbol{\theta} \gets \mathcal{L}_{\text{rep}} $ (Eq.~(\ref{eq:loss}))
    
\ENDFOR
\end{algorithmic}
\end{algorithm}

\subsection{Training Algorithm Flow}
To implement and clarify our method, we provide the algorithm flow for training in Algorithm \ref{alg:promrl}. Building upon our theoretical derivation, our method is straightforward, applying a few steps to update the parameters in batches of data. Observed unimodal content will be encoded and then used to determine the posterior, which also makes the generative parameters resolvable within this multimodal learning framework.   

  \begin{table*}[t]
\centering
\caption{\textbf{Multimodal retrieval results~(\%) in terms of Recall@1} on video-text (T$\rightarrow$V and V$\rightarrow$T) and audio-text (T$\rightarrow$A and A$\rightarrow$T) datasets. ``$\uparrow$'' indicates the model continually trained with missing modality datasets. The best result in each case is marked in bold, and the second-best result is underlined. Increment points are computed compared with VAST. }
\vspace{-2mm}
\label{tab:overall_performance}
\begin{adjustbox}{width=\textwidth,center}
\begin{tabular}{@{}l|ll|ll|ll|ll|ll|ll|c@{}}
\toprule
 & \multicolumn{2}{c|}{MSR-VTT} & \multicolumn{2}{c|}{DiDeMo} & \multicolumn{2}{c|}{ActivityNet}  & \multicolumn{2}{c|}{VATEX}   & \multicolumn{2}{c|}{AudioCaps}  & \multicolumn{2}{c|}{Clotho} & Avg. \\  & T$\rightarrow$V    & V$\rightarrow$T   & T$\rightarrow$V     & V$\rightarrow$T  & T$\rightarrow$V    & V$\rightarrow$T  & T$\rightarrow$V    & V$\rightarrow$T  & T$\rightarrow$A    & A$\rightarrow$T   & T$\rightarrow$A    & A$\rightarrow$T    \\ \midrule 
ImageBind  & 36.8    &  -  & -        & -     & -    & -  & -&-&9.3 &- & 6.0 & - & -\\
InternVideo-L & 40.7 &    39.6   & 31.5     & 33.5     & 30.7      & 31.4 & 49.5 & 69.5 &- &- & - & -& -\\
LanguageBind & 44.8     & 40.9   & 39.9    & 39.8  & 41.0      &  39.1 & - & - &19.7 &- & 16.7 & - & -\\ \midrule 
 VAST       & 50.5                                                 & 49.0                                                 & 48.6                                                 & 46.9                                                 & 51.7                                                 & 48.8                                                 & 75.9                                                 & 74.8                                                 & 33.7                                                 & 32.2                                                 & 12.4                                                 & 13.0   & 44.8                                              \\
 GRAM       & 52.1\textcolor{lightgreen}{$^\text{+1.6}$}           & 51.8\textcolor{lightgreen}{$^\text{+2.8}$}           & 53.1\textcolor{lightgreen}{$^\text{+4.5}$}           & 50.7\textcolor{lightgreen}{$^\text{+3.8}$}           & 54.5\textcolor{lightgreen}{$^\text{+2.8}$}           & 48.3\textcolor{lightgreen}{$^\text{-0.5}$}           & 77.5\textcolor{lightgreen}{$^\text{+1.6}$}           & 74.7\textcolor{lightgreen}{$^\text{-0.1}$}           & 34.6\textcolor{lightgreen}{$^\text{+0.9}$}           & 35.2\textcolor{lightgreen}{$^\text{+3.0}$}           & 15.9\textcolor{lightgreen}{$^\text{+3.5}$}           & 16.2\textcolor{lightgreen}{$^\text{+3.2}$}         & 47.1   \\
 TRIANGLE   & 54.3\textcolor{lightgreen}{$^\text{+3.8}$}           & 51.7\textcolor{lightgreen}{$^\text{+2.7}$}           & 53.4\textcolor{lightgreen}{$^\text{+4.8}$}           & 52.7\textcolor{lightgreen}{$^\text{+5.8}$}           & 55.4\textcolor{lightgreen}{$^\text{+3.7}$}           & 50.9\textcolor{lightgreen}{$^\text{+2.1}$}           & 79.9\textcolor{lightgreen}{$^\text{+4.0}$}           & 74.8\textcolor{lightgreen}{$^\text{+0.0}$}           & 37.2\textcolor{lightgreen}{$^\text{+3.5}$}           & 37.2\textcolor{lightgreen}{$^\text{+5.0}$}           & 15.3\textcolor{lightgreen}{$^\text{+2.9}$}           & 13.7\textcolor{lightgreen}{$^\text{+0.7}$}      & 48.0     \\
 PMRL       & 55.1\textcolor{lightgreen}{$^\text{+4.6}$}           & 53.5\textcolor{lightgreen}{$^\text{+4.5}$}           & 53.5\textcolor{lightgreen}{$^\text{+4.9}$}           & 51.3\textcolor{lightgreen}{$^\text{+4.4}$}           & 56.0\textcolor{lightgreen}{$^\text{+4.3}$}           & 49.6\textcolor{lightgreen}{$^\text{+0.8}$}           & 80.5\textcolor{lightgreen}{$^\text{+4.6}$}           & 75.2\textcolor{lightgreen}{$^\text{+0.4}$}           & 36.1\textcolor{lightgreen}{$^\text{+2.4}$}           & 33.9\textcolor{lightgreen}{$^\text{+1.7}$}           & 16.8\textcolor{lightgreen}{$^\text{+4.4}$}           & 16.1\textcolor{lightgreen}{$^\text{+3.1}$}      & 48.1        \\
 \midrule
VAST$\uparrow$        & 58.5\textcolor{lightgreen}{$^\text{+8.0}$}            & \underline{60.2}\textcolor{lightgreen}{$^\text{+11.2}$}           & 53.9\textcolor{lightgreen}{$^\text{+5.3}$}           & 53.1\textcolor{lightgreen}{$^\text{+6.2}$}           & 55.7\textcolor{lightgreen}{$^\text{+4.0}$}           & \underline{53.9}\textcolor{lightgreen}{$^\text{+5.1}$}           & 80.0\textcolor{lightgreen}{$^\text{+4.1}$}           & 77.9\textcolor{lightgreen}{$^\text{+3.1}$}           & 49.1\textcolor{lightgreen}{$^\text{+15.4}$}           & \textbf{53.3\textcolor{lightgreen}{$^\text{+21.1}$}}           & 21.8\textcolor{lightgreen}{$^\text{+9.4}$}            & 21.8\textcolor{lightgreen}{$^\text{+8.8}$}     & 53.3       \\
GRAM$\uparrow$  & 59.7\textcolor{lightgreen}{$^\text{+9.2}$}  & 57.2\textcolor{lightgreen}{$^\text{+8.2}$}  & 54.8\textcolor{lightgreen}{$^\text{+6.2}$} & 53.1\textcolor{lightgreen}{$^\text{+6.2}$} & \underline{56.2}\textcolor{lightgreen}{$^\text{+4.5}$} & 53.5\textcolor{lightgreen}{$^\text{+4.7}$} & \underline{80.5}\textcolor{lightgreen}{$^\text{+4.6}$} & \textbf{79.2\textcolor{lightgreen}{$^\text{+4.4}$}} & 49.1\textcolor{lightgreen}{$^\text{+15.4}$} & 51.7\textcolor{lightgreen}{$^\text{+19.5}$} & 20.6\textcolor{lightgreen}{$^\text{+8.2}$} & 19.5\textcolor{lightgreen}{$^\text{+6.5}$}  & 52.9 \\
TRIANGLE$\uparrow$  & 57.6\textcolor{lightgreen}{$^\text{+7.1}$}           & 58.4\textcolor{lightgreen}{$^\text{+9.4}$}           & 51.7\textcolor{lightgreen}{$^\text{+3.1}$}           & 51.1\textcolor{lightgreen}{$^\text{+4.2}$}           & 54.2\textcolor{lightgreen}{$^\text{+2.5}$}           & 51.0\textcolor{lightgreen}{$^\text{+2.2}$}           & 77.9\textcolor{lightgreen}{$^\text{+2.0}$}           & 76.6\textcolor{lightgreen}{$^\text{+1.8}$}           & 48.3\textcolor{lightgreen}{$^\text{+14.6}$}          & 51.7\textcolor{lightgreen}{$^\text{+19.5}$}          & 19.9\textcolor{lightgreen}{$^\text{+7.5}$}           & 20.2\textcolor{lightgreen}{$^\text{+7.2}$}    & 51.6        \\
PMRL$\uparrow$   & \underline{60.1}\textcolor{lightgreen}{$^\text{+9.6}$}           & 59.2\textcolor{lightgreen}{$^\text{+10.2}$}          & \underline{55.1}\textcolor{lightgreen}{$^\text{+6.5}$}           & \underline{53.3}\textcolor{lightgreen}{$^\text{+6.4}$}           & 55.8\textcolor{lightgreen}{$^\text{+4.1}$}           & \textbf{54.0\textcolor{lightgreen}{$^\text{+5.2}$}}           & 80.4\textcolor{lightgreen}{$^\text{+4.5}$}           & \underline{78.7}\textcolor{lightgreen}{$^\text{+3.9}$}           & \textbf{50.4\textcolor{lightgreen}{$^\text{+16.7}$}}          & \underline{52.0}\textcolor{lightgreen}{$^\text{+19.8}$}          & \underline{23.5}\textcolor{lightgreen}{$^\text{+11.1}$}          & \textbf{23.1\textcolor{lightgreen}{$^\text{+10.1}$}}     & \underline{53.8} \\

  \midrule
 \textbf{CalMRL$\uparrow$}    & \textbf{61.1\textcolor{darkgreen}{$^\textbf{+10.6}$}} & \textbf{61.1\textcolor{darkgreen}{$^\textbf{+12.1}$}} & \textbf{55.4\textcolor{darkgreen}{$^\textbf{+6.8}$}} & \textbf{53.7\textcolor{darkgreen}{$^\textbf{+6.8}$}} & \textbf{57.1\textcolor{darkgreen}{$^\textbf{+5.4}$}} & 53.6\textcolor{darkgreen}{$^\textbf{+4.8}$} & \textbf{81.3\textcolor{darkgreen}{$^\textbf{+5.4}$}} & \textbf{79.2\textcolor{darkgreen}{$^\textbf{+4.4}$}} & \underline{50.1}\textcolor{darkgreen}{$^\textbf{+16.4}$} & {51.0}\textcolor{darkgreen}{$^\textbf{+18.8}$} & \textbf{23.8\textcolor{darkgreen}{$^\textbf{+11.4}$}} & \underline{22.4}\textcolor{darkgreen}{$^\textbf{+9.4}$} & \textbf{54.2} \\

 \bottomrule
\end{tabular}
\end{adjustbox}
\end{table*}

\begin{table}
\centering
\caption{\textbf{Multimodal classification results~(\%) in terms of Accuracy} on video-text and audio-text datasets across different models.}
\vspace{-2mm}
\label{tab:overall_performance_class}
\begin{adjustbox}{width=\linewidth,center}
\begin{tabular}{l|llll|l}
\toprule
 & VGGSound & UCF101 & AudioSet & ESC50 & Avg. \\
\midrule
ImageBind & 25.34 & 69.31 & 14.92 & 58.74 & 42.08 \\
\midrule
VAST$\uparrow$ & 25.98 & 74.55 & 15.03 & 59.99 & 43.89 \\
GRAM$\uparrow$ & 25.78 & 73.56 & 15.65 & 58.50 & 43.37 \\
TRIANGLE$\uparrow$ & 25.69 & 73.86 & 15.20 & 59.24 & 43.50 \\
PMRL$\uparrow$ & 25.24 & 77.13 & 15.29 & 58.50 & 44.04 \\
\midrule
\textbf{CalMRL$\uparrow$} & \textbf{26.09}\textcolor{darkgreen}{$^\textbf{+0.85}$} & \textbf{78.91}\textcolor{darkgreen}{$^\textbf{+1.78}$} & \textbf{16.01}\textcolor{darkgreen}{$^\textbf{+0.72}$} & \textbf{59.75}\textcolor{darkgreen}{$^\textbf{+1.25}$} & \textbf{45.19}\textcolor{darkgreen}{$^\textbf{+1.15}$} \\
\bottomrule
\end{tabular}
\end{adjustbox}
\vspace{-6mm}
\end{table}

\section{Experiments}

In this section, we conduct experiments to address the following research question:
\begin{itemize}[leftmargin=*]
    \item \textbf{RQ1:} Does CalMRL outperform other multimodal representation methods under the missing-modality setting? 
    \item \textbf{RQ2:} What is the contribution of CalMRL to different missing modalities? Does it satisfy the theoretical insight to calibrate for a better alignment with empirical results?
    \item \textbf{RQ3:} How is the training of CalMRL stable? Can it maintain the distribution of multimodal representations?
\end{itemize}

\subsection{Experimental Setup}

\noindent\textbf{Datasets.} We first adopt the VAST-150K with complete modalities for warming up parameters, following \cite{chen2023vast, cicchetti2024gramian}. Then we train the model on datasets with missing modality. Specifically, we select two datasets: MSRVTT \cite{bain2021frozen} for vision-text pairs and AudioCaps \cite{kim2019audiocaps} for audio-text pairs, covering two mainstream types of multimodal datasets (\textit{i.e.}, V$\leftrightarrow$T, A$\leftrightarrow$T). 
To evaluate our method, we select 10 benchmarks, including MSR-VTT \cite{chen2011collecting}, DiDeMo \cite{anne2017localizing}, ActivityNet \cite{krishna2017dense}, and VATEX \cite{wang2019vatex} for vision-text retrieval, AudioCaps \cite{kim2019audiocaps}, and Clotho \cite{drossos2020clotho} for audio-text retrieval.  Only test splits are used for evaluation to avoid any information leakage. We also evaluate our method on the multimodal classification task via fine-tuning on VGGSound~\cite{chen2020vggsound} and UCF101~\cite{soomro2012ucf101}, and evaluating on four testing datasets, including VGGSound, UCF101, AudioSet~\cite{gemmeke2017audio}, and ESC50~\cite{piczak2015esc}, in Section~\ref{sec:overall_performance}. We detail the statistics of datasets in Appendix~\ref{sec:appendix:datasets}.

\noindent\textbf{Baselines \& evaluation metrics.} We compare our method with existing well-trained models, including ImageBind \cite{girdhar2023imagebind}, InternVideo \cite{wang2022internvideo}, LanguageBind \cite{zhu2023languagebind}, VAST \cite{chen2023vast}, GRAM \cite{cicchetti2024gramian}, TRIANGLE \cite{giordano2025triangle}, and PMRL \cite{liu2026principled}. We detail the implementation of baselines, especially their adaptation for missing modalities in Appendix~\ref{sec:appendix:adaption_baselines}. 
Performance is evaluated using Recall@1 for retrieval and Accuracy for classification. 
Detailed results with different top-k ($\{5,10\}$) on retrieval tasks are shown in Appendix~\ref{sec:appendix:addtional_results}, with results reported for bidirectional retrieval (\textit{e.g.}, T$\rightarrow$V and V$\rightarrow$T).

\noindent\textbf{Implementation details.} We implement our model upon VAST \cite{chen2023vast} which supports four modalities, \textit{i.e.}, vision, caption, audio, and subtitle. We also efficiently adapt different multimodal alignment methods on ImageBind evaluate its classification performance. The detailed model architecture is in Appendix~\ref{sec:appendix:model_architecture}. VAST, GRAM, TRIANGLE, and PMRL are all fine-tuned on the same training datasets aligning with the missing modality training to ensure a fair comparison with CalMRL. We also report their base performance only trained on the full modality dataset (VAST) in our empirical results (\textit{cf.}, Section \ref{sec:further_analysis}).

\begin{table*}[t]
\centering
\caption{\textbf{Multimodal retrieval results~(\%) for models trained on sole datasets.} ``$\uparrow^{\text{VT}}$'' and ``$\uparrow^{\text{AT}}$'' indicate the model continually trained with video-text and audio-text modality datasets, respectively. The best result in each case is marked in bold, and the second-best result is underlined. Increment points are computed compared with VAST. }
\vspace{-2mm}
\label{tab:sole_datasets}
\begin{adjustbox}{width=\textwidth,center}
\begin{tabular}{@{}l|ll|ll|ll|ll|ll|ll|c@{}}
\toprule
 & \multicolumn{2}{c|}{MSR-VTT} & \multicolumn{2}{c|}{DiDeMo} & \multicolumn{2}{c|}{ActivityNet}  & \multicolumn{2}{c|}{VATEX}   & \multicolumn{2}{c|}{AudioCaps}  & \multicolumn{2}{c}{Clotho} & Avg.  \\  & T$\rightarrow$V    & V$\rightarrow$T   & T$\rightarrow$V     & V$\rightarrow$T  & T$\rightarrow$V    & V$\rightarrow$T  & T$\rightarrow$V    & V$\rightarrow$T  & T$\rightarrow$A    & A$\rightarrow$T   & T$\rightarrow$A    & A$\rightarrow$T    \\ \midrule 
VAST$\uparrow^{\text{AT}}$ & 53.1\textcolor{lightgreen}{$^\text{+2.6}$}            & 50.9\textcolor{lightgreen}{$^\text{+1.9}$}            & 45.0\textcolor{lightgreen}{$^\text{-3.6}$}           & 46.0\textcolor{lightgreen}{$^\text{-0.9}$}           & 48.8\textcolor{lightgreen}{$^\text{-2.9}$}           & 48.5\textcolor{lightgreen}{$^\text{-0.3}$}           & 77.3\textcolor{lightgreen}{$^\text{+1.4}$}           & 75.9\textcolor{lightgreen}{$^\text{+1.1}$}           & 51.1\textcolor{lightgreen}{$^\text{+17.4}$}           & 52.1\textcolor{lightgreen}{$^\text{+19.9}$}           & 21.3\textcolor{lightgreen}{$^\text{+8.9}$}           & 21.1\textcolor{lightgreen}{$^\text{+8.1}$}       & 49.3    \\
GRAM$\uparrow^{\text{AT}}$   & 49.0\textcolor{lightgreen}{$^\text{-1.5}$}           & 49.3\textcolor{lightgreen}{$^\text{+0.3}$}           & 48.5\textcolor{lightgreen}{$^\text{-0.1}$}           & 48.3\textcolor{lightgreen}{$^\text{+1.4}$}           & 49.2\textcolor{lightgreen}{$^\text{-2.5}$}           & 48.0\textcolor{lightgreen}{$^\text{-0.8}$}           & 58.1\textcolor{lightgreen}{$^\text{-17.8}$}          & 74.1\textcolor{lightgreen}{$^\text{-0.7}$}           & \underline{53.0}\textcolor{lightgreen}{$^\text{+19.3}$}          & 52.1\textcolor{lightgreen}{$^\text{+19.9}$}          & \underline{22.6}\textcolor{lightgreen}{$^\text{+10.2}$}          & \textbf{22.5\textcolor{lightgreen}{$^\text{+9.5}$}}    & 47.9       \\
TRIANGLE$\uparrow^{\text{AT}}$ & 55.8\textcolor{lightgreen}{$^\text{+5.3}$}           & 52.4\textcolor{lightgreen}{$^\text{+3.4}$}           & \underline{50.1}\textcolor{lightgreen}{$^\text{+1.5}$}           & 48.9\textcolor{lightgreen}{$^\text{+2.0}$}           & 50.2\textcolor{lightgreen}{$^\text{-1.5}$}           & \textbf{50.0\textcolor{lightgreen}{$^\text{+1.2}$}}          & \textbf{79.8\textcolor{lightgreen}{$^\text{+3.9}$}}          & \underline{76.1}\textcolor{lightgreen}{$^\text{+1.3}$}           & 47.0\textcolor{lightgreen}{$^\text{+13.3}$}          & 50.9\textcolor{lightgreen}{$^\text{+18.7}$}          & 16.6\textcolor{lightgreen}{$^\text{+4.2}$}           & 18.4\textcolor{lightgreen}{$^\text{+5.4}$}        & 49.7   \\
PMRL$\uparrow^{\text{AT}}$   & \underline{56.2}\textcolor{lightgreen}{$^\text{+5.7}$}           & \underline{52.7}\textcolor{lightgreen}{$^\text{+3.7}$}           & 48.7\textcolor{lightgreen}{$^\text{+0.1}$}           & \underline{49.4}\textcolor{lightgreen}{$^\text{+2.5}$}           & \underline{52.0}\textcolor{lightgreen}{$^\text{+0.3}$}           & 49.3\textcolor{lightgreen}{$^\text{+0.5}$}           & 78.8\textcolor{lightgreen}{$^\text{+2.9}$}           & \textbf{76.6\textcolor{lightgreen}{$^\text{+1.8}$}}           & {52.0}\textcolor{lightgreen}{$^\text{+18.3}$}          & \underline{54.0}\textcolor{lightgreen}{$^\text{+21.8}$}          & \textbf{22.7\textcolor{lightgreen}{$^\text{+10.3}$} }         & \underline{21.8}\textcolor{lightgreen}{$^\text{+8.8}$}       & \underline{51.2}    \\
\midrule
\textbf{CalMRL$\uparrow^{\text{AT}}$}    &  \textbf{56.4\textcolor{darkgreen}{$^\text{+5.9}$} }           & \textbf{53.3\textcolor{darkgreen}{$^\text{+4.3}$}   }         & \textbf{50.5\textcolor{darkgreen}{$^\text{+1.9}$}}           & \textbf{50.7\textcolor{darkgreen}{$^\text{+3.8}$}}           & \textbf{53.8\textcolor{darkgreen}{$^\text{+2.1}$}}           & \underline{49.8}\textcolor{darkgreen}{$^\text{+1.0}$}           & \underline{79.2}\textcolor{darkgreen}{$^\text{+3.3}$}           & \textbf{76.6\textcolor{darkgreen}{$^\text{+1.8}$}}           & \textbf{53.1\textcolor{darkgreen}{$^\text{+19.4}$}}           & \textbf{54.1\textcolor{darkgreen}{$^\text{+21.9}$} }          & 21.3\textcolor{darkgreen}{$^\text{+8.9}$}           & 21.6\textcolor{darkgreen}{$^\text{+8.6}$}        & \textbf{51.7}  \\
\bottomrule\toprule
VAST$\uparrow^{\text{VT}}$  & 59.5\textcolor{lightgreen}{$^\text{+9.0}$}            & \textbf{60.3\textcolor{lightgreen}{$^\text{+11.3}$}}           & 54.6\textcolor{lightgreen}{$^\text{+6.0}$}           & \textbf{55.1\textcolor{lightgreen}{$^\text{+8.2}$}}           & \textbf{57.6\textcolor{lightgreen}{$^\text{+5.9}$}}           & \textbf{54.9}\textcolor{lightgreen}{$^\text{+6.1}$}           & \underline{80.1}\textcolor{lightgreen}{$^\text{+4.2}$}           & \underline{78.3}\textcolor{lightgreen}{$^\text{+3.5}$}           & 31.5\textcolor{lightgreen}{$^\text{-2.2}$}            & 33.4\textcolor{lightgreen}{$^\text{+1.2}$}            & 15.5\textcolor{lightgreen}{$^\text{+3.1}$}           & 15.5\textcolor{lightgreen}{$^\text{+2.5}$}           & 49.7\\
GRAM$\uparrow^{\text{VT}}$    & 60.1\textcolor{lightgreen}{$^\text{+9.6}$}            & 57.8\textcolor{lightgreen}{$^\text{+8.8}$}            & \underline{56.5}\textcolor{lightgreen}{$^\text{+7.9}$}           & 54.0\textcolor{lightgreen}{$^\text{+7.1}$}           & 56.2\textcolor{lightgreen}{$^\text{+4.5}$}           & \underline{53.5}\textcolor{lightgreen}{$^\text{+4.7}$}           & 78.9\textcolor{lightgreen}{$^\text{+3.0}$}           & \underline{78.3}\textcolor{lightgreen}{$^\text{+3.5}$}           & \textbf{32.5\textcolor{lightgreen}{$^\text{-1.2}$}}           & \textbf{36.8\textcolor{lightgreen}{$^\text{+4.6}$}}            & 16.1\textcolor{lightgreen}{$^\text{+3.7}$}            & 15.2\textcolor{lightgreen}{$^\text{+2.2}$}          & 49.7 \\
 TRIANGLE$\uparrow^{\text{VT}}$ & 57.8\textcolor{lightgreen}{$^\text{+7.3}$}           & 58.7\textcolor{lightgreen}{$^\text{+9.7}$}           & 53.1\textcolor{lightgreen}{$^\text{+4.5}$}           & 50.9\textcolor{lightgreen}{$^\text{+4.0}$}           & 56.3\textcolor{lightgreen}{$^\text{+4.6}$}           & 51.6\textcolor{lightgreen}{$^\text{+2.8}$}           & 78.4\textcolor{lightgreen}{$^\text{+2.5}$}           & 77.3\textcolor{lightgreen}{$^\text{+2.5}$}           & 31.2\textcolor{lightgreen}{$^\text{-2.5}$}           & 32.5\textcolor{lightgreen}{$^\text{+0.3}$}           & 16.3\textcolor{lightgreen}{$^\text{+3.9}$}           & 14.7\textcolor{lightgreen}{$^\text{+1.7}$}           & 48.2 \\
PMRL$\uparrow^{\text{VT}}$     & \underline{60.7}\textcolor{lightgreen}{$^\text{+10.2}$}          & \underline{60.0}\textcolor{lightgreen}{$^\text{+11.0}$}          & 56.1\textcolor{lightgreen}{$^\text{+7.5}$}           & 53.5\textcolor{lightgreen}{$^\text{+6.6}$}           & \underline{57.5}\textcolor{lightgreen}{$^\text{+5.8}$}           & 53.2\textcolor{lightgreen}{$^\text{+4.4}$}           & 79.7\textcolor{lightgreen}{$^\text{+3.8}$}           & 78.1\textcolor{lightgreen}{$^\text{+3.3}$}           & \underline{32.0}\textcolor{lightgreen}{$^\text{-1.7}$}           & \underline{34.8}\textcolor{lightgreen}{$^\text{+2.6}$}           & \underline{18.2}\textcolor{lightgreen}{$^\text{+5.8}$}           & \textbf{15.9}\textcolor{lightgreen}{$^\text{+2.9}$}         & \underline{50.0}  \\
 \midrule
\textbf{CalMRL$\uparrow^{\text{VT}}$}   & \textbf{61.1\textcolor{lightgreen}{$^\text{+10.6}$}}           & \textbf{60.3\textcolor{lightgreen}{$^\text{+11.3}$}}          & \textbf{57.5\textcolor{lightgreen}{$^\text{+8.9}$}}           & \underline{54.4}\textcolor{lightgreen}{$^\text{+7.5}$}           & \underline{57.5}\textcolor{lightgreen}{$^\text{+5.8}$}           & 52.2\textcolor{lightgreen}{$^\text{+3.4}$}           & \textbf{81.5\textcolor{lightgreen}{$^\text{+5.6}$}}           & \textbf{79.4\textcolor{lightgreen}{$^\text{+4.6}$}}           & 31.4\textcolor{lightgreen}{$^\text{-2.3}$}            & \underline{34.8}\textcolor{lightgreen}{$^\text{+2.6}$}            & \textbf{18.9\textcolor{lightgreen}{$^\text{+6.5}$}}           & \underline{15.8}\textcolor{lightgreen}{$^\text{+2.8}$}       & \textbf{50.4}    \\
 \bottomrule
\end{tabular}
\end{adjustbox}
\end{table*}

\begin{figure*}[t]
    \centering
    \begin{minipage}[t]{0.312\linewidth}
        \centering
\includegraphics[width=\textwidth, clip, trim=0mm 0mm 0 0mm]{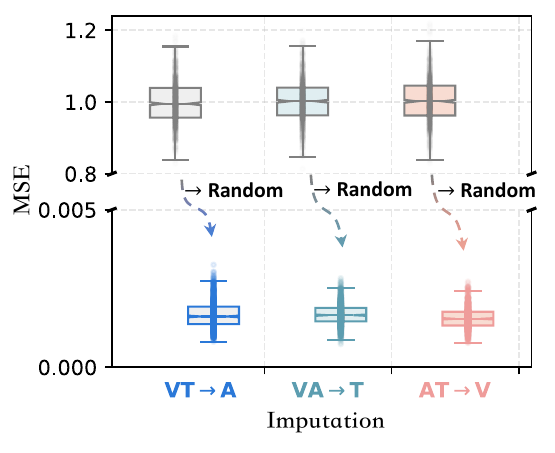} 
         \vspace{-7mm}
        \caption{\textbf{MSEs between real and imputed representations.} ``$\rightarrow$'' marks the direction of imputation; \textbf{Random} refers to representations drawn at random.}
        \label{fig:mse}
    \end{minipage}\hspace{6pt}
    \begin{minipage}[t]{0.32\linewidth}
        \centering
\includegraphics[width=\textwidth, clip, trim=0mm 0mm 0 2mm]{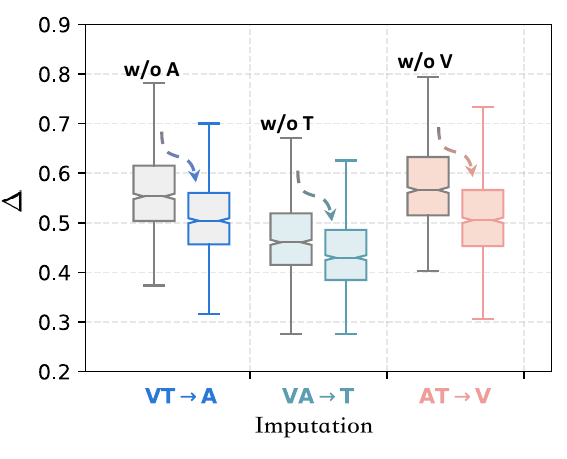}
         \vspace{-7mm}
        \caption{\textbf{Comparison of anchor shift ($\Delta$) before and after calibration.} The left box with a gray border shows the anchor shift with missing modalities (w/o).}
        \label{fig:anchor}
    \end{minipage}\hspace{6pt}
    \begin{minipage}[t]{0.32\linewidth}
        \centering
\includegraphics[width=\textwidth, clip, trim=0mm -5mm 0 0mm]{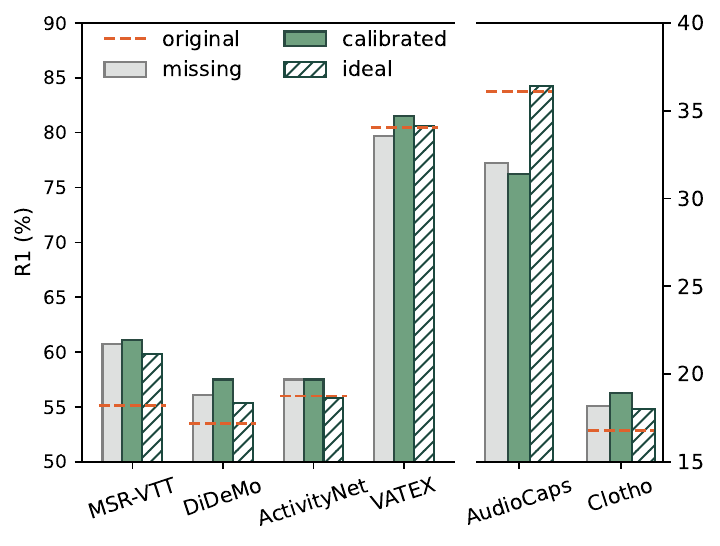}
         \vspace{-7mm}
        \caption{\textbf{The performance comparison across missing, calibrated, and full (``ideal'') modalities.} 
        All the models are trained on MSR-VTT. 
        }
        \label{fig:ideal_comparison}
    \end{minipage}
\end{figure*}

\subsection{Performance Comparison (RQ1)}
\label{sec:overall_performance}
We report the results on ten datasets, covering two main multimodal retrieval and classification tasks, as shown in Tables \ref{tab:overall_performance} and \ref{tab:overall_performance_class}. All the models are trained on a complete-modality dataset initially, and ``$\uparrow$'' denotes the continual training on missing modality datasets. Accordingly, we draw the following conclusions: 

\ding{172} \textit{Learning with complete-modality simultaneously achieves higher performance.} Initial models, \textit{e.g.}, ImageBind, InternVideo-L, and LanguageBind, extend the pairwise contrastive learning paradigm. However, they generally perform worse than models trained on datasets where all modalities are available simultaneously. New learning objectives have been introduced specifically to optimize multiple modalities concurrently. Among these methods, PMRL demonstrates relatively better performance. The aligned modalities naturally reflect different aspects of a single common instance. This inherent alignment intuitively provides significant benefits for multimodal representation learning.

\ding{173} \textit{Missing modalities can boost the model yet incrementally.} 
When certain modalities are missing, virtually all models demonstrate performance improvements, particularly on in-domain datasets (\textit{i.e.}, MSR-VTT and AudioCaps). Observable performance gains are also generalized to various out-of-domain datasets, such as DiDeMo and ActivityNet, though these gains are incremental. The performance boost on audio-text datasets (\textit{i.e.}, AudioCaps and Clotho) is significantly greater compared to the boost observed on vision-text alignment tasks.
This disparity indicates that vision-text alignment has benefited from extensive prior research and advancements, whereas audio-text alignment appears to have substantially greater room for improvement (corresponds to the visualization in Figure \ref{fig:tsne}).

\ding{174} \textit{CalMRL further improves without incorporating new information.} 
Among all the methods, CaLMRL demonstrates superior performance improvements over its backbone model, VAST, and surpasses the SOTA methods in most cases. Rather than requiring new datasets, CaLMRL adapts existing bimodal datasets to enhance the model using a simple generative approach, showcasing strong promise.

\ding{175} \textit{Further classification results also reflect similar effects.} In general, CalMRL can also achieve consistent performance gains across diverse multimodal classification benchmarks, including UCF101, VGGSound, AudioSet, and ESC50 tasks. As shown in Table \ref{tab:overall_performance_class}, CalMRL outperforms the baselines, reaching an average accuracy of 45.19\%. Notably, it achieves a significant improvement of +1.78\% on UCF101 tuning with the same strategy as PMRL. These results further validate that the representation alignment and generative imputation in CalMRL effectively enhance the model's robustness and generalization capabilities in missing-modality scenarios.

\subsection{Further Analysis (R2 \& R3)}
\label{sec:further_analysis}

\begin{figure*}
    \centering
    \includegraphics[width=\linewidth]{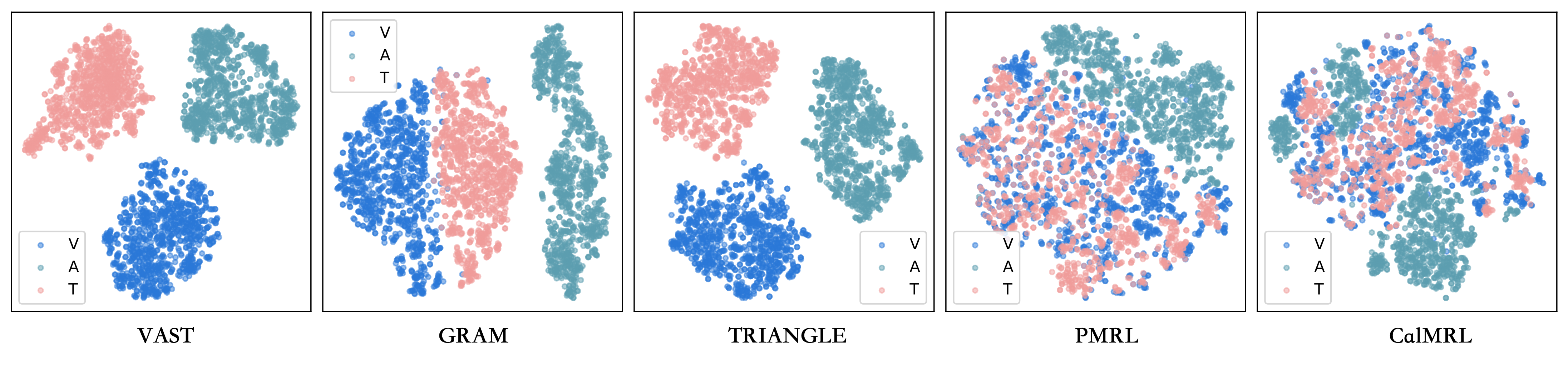}
    \vspace{-6mm}
    \caption{\textbf{t-SNE visualization on multimodal representations generated by different models.} Existing models, under missing modality training, present clearly separated clusters for each modality (distinct modal boundaries). Fortunately, CalMRL mitigates this issue.}
    \vspace{-3mm}
    \label{fig:tsne}
\end{figure*}

\noindent\textbf{Sole datasets with calibration (R2).} To unravel the effect of different missing modality datasets, we conduct experiments on a single dataset with calibration. Table \ref{tab:sole_datasets} showcases the performance evaluated on different variants of models. ``$\uparrow^{\text{AT}}$'' indicates the training with only audio-text dataset (\textit{i.e.}, AudioCaps). The results for all given benchmarks are reported, and we have the following insights. 
\ding{172} \textit{Training on the sole dataset can improve the relevant capabilities of modalities while harming others.} 
For instance, training on an audio-text dataset, we can observe a significant improvement on AudioCpas and Clotho (out-of-domain), and a performance drop on other video-text datasets, and vice versa. Such harmfulness can be found clearly under only the vision-text dataset training. 
\ding{173} \textit{CalMRL heals to some extent and even sometimes achieves better performance.} In audio-text training, CalMRL resists performance degradation in most vision-text benchmarks, and even gains better results in AudioCaps. Combining the results in Table~\ref{tab:overall_performance}, CalMRL demonstrates the collective benefit of training on a mixture of datasets, resulting in improved performance.
These results provide strong evidence to shed light on how CalMRL works: calibrating the alignment to resist the degradation for missing modalities to maintain an overall better performance. 

\noindent\textbf{Compensating for the anchor shift (R2).} We measure the anchor shift before and after calibration to validate its effectiveness. Figures \ref{fig:mse} and \ref{fig:anchor} illustrate that \ding{172} \textit{our method can effectively reconstruct the original representations} and \ding{173} \textit{the anchor shift is mitigated as expected}.
This indicates that our method, even with a simple generative model, is capable of learning multimodal connections at the representation level. Without introducing any new information, the incomplete alignment can be approximated to the complete one via the mitigated anchor shift.

\begin{figure}
    \centering
    \includegraphics[width=\linewidth]{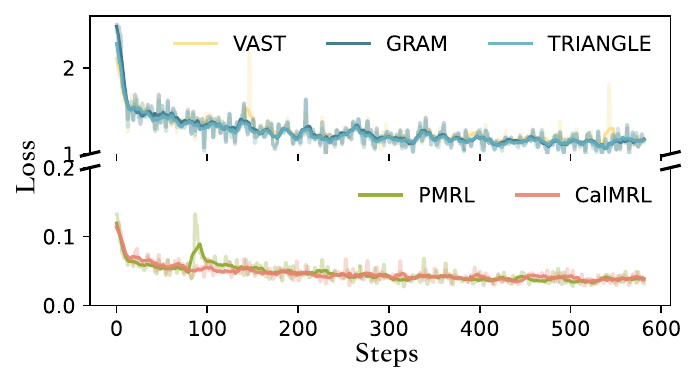}
    \vspace{-6mm}
    \caption{{Loss curves across models on the training phase.} }
    \label{fig:loss}
    \vspace{-6mm}
\end{figure}

\noindent\textbf{Approaching to the \textit{``ideal''} anchor (R2).} We conduct experiments where the model was trained on a dataset with synthetic complete modalities to mimic the complete-modality scenario. Figure \ref{fig:ideal_comparison} shows the comparison of the variant without calibration and the one with complete modalities on the MSR-VTT. Compared to the original performance, training on complete modalities can best maintain the alignment, achieving results around or above the red line.
CalMRL outperforms the missing-modality variant in most cases and even the ``ideal'' (complete-modality training) in the VATEX dataset. We attribute it to the mitigation of the inherent noise in the full datasets. In contrast, imposing learning on un-match examples leads to certain negative effects. 

\noindent\textbf{Training stability (R3).} 
We plot the curves of the training loss $\mathcal{L}_{\text{rep}}$ for different models in Figure \ref{fig:loss}. VAST, GRAM, and TRIANGLE all utilize text-anchored contrastive learning, which leads to relatively larger losses, while CalMRL follows PMRL, optimizing with smaller ones. Overall, we can observe that the loss curve of CalMRL is more stable and exhibits a steadier decreasing trend compared to the others. Despite employing bi-step learning for missing modalities, CalMRL is capable of managing the training and achieving convergence.

\noindent\textbf{Case visualization (R3).} 
To intuitively understand the inner workings of CalMRL, we visualize the representations as points within a 2D plane to analyze their distributions. As shown in Figure \ref{fig:tsne}, VAST, GRAM, and TRIANGLE demonstrate distinct, well-separated regions for each modality. While PMRL successfully aligns the vision and text representations, it excludes the audio representation with a clear dividing line. In contrast, CalMRL alleviates this separation, pulling the multiple modalities closer together. We attribute this improvement to the alignment calibration, the core design philosophy of CalMRL, which effectively prevents the segregation of different modal representations.

  \section{Conclusion}
In this paper, we presented CalMRL, a calibrated multimodal representation learning framework for handling missing modalities. We theoretically revealed the anchor shift phenomenon and proposed a generative imputation mechanism to reconstruct representations and calibrate alignment. Through a bi-step optimization with closed-form inference and iterative refinement, CalMRL achieves stable convergence and strong compatibility with existing alignment paradigms. Experiments across diverse benchmarks verify its superiority. 
Looking ahead, future explorations can extend our method to broader applications and focus on: calibrating the shift without imputation, incorporating more datasets to enhance multimodal models, and interpreting multimodal connections to inspire further advancement.

\section*{Impact Statement}
This paper presents work whose goal is to advance the field of multimodal learning by addressing the ``anchor shift'' problem in modality missing scenarios. By providing a theoretical and practical framework for handling missing modalities, our work facilitates the development of more flexible AI systems that can operate in data-constrained environments. While the use of generative models to impute missing data carries the general risk of perpetuating biases inherent in the training distribution, we believe there are no specific societal or ethical consequences of our work that must be uniquely highlighted here beyond the standard considerations for the field.

\section*{Acknowledgement}
This research/project is supported by the National Research Foundation, Singapore under its National Large Language Models Funding Initiative (AISG Award No: AISG-NMLP-2024-002). Any opinions, findings and conclusions or recommendations expressed in this material are those of the author(s) and do not reflect the views of National Research Foundation, Singapore.

Jiaheng Wei is partially supported by Guangdong Provincial Key Lab of Integrated Communication, Sensing and Computation for Ubiquitous Internet of Things (No. 2023B1212010007). Shuo Yang is supported by the Shenzhen Fundamental Research Program (JCYJ20250604145514018), Guangdong Basic and Applied Basic Research Foundation (General Program, No. 2026A1515011557), and the NSFC Young Scientists Fund (No. 62506096). Xiu Su is supported by the National Natural Science Foundation of China (No. 62406347).

  \bibliography{reference}

@inproceedings{cai2018deep,
  title={Deep adversarial learning for multi-modality missing data completion},
  author={Cai, Lei and Wang, Zhengyang and Gao, Hongyang and Shen, Dinggang and Ji, Shuiwang},
  booktitle={KDD},
  pages={1158--1166},
  year={2018}
}

@article{wu2024deep,
  title={Deep multimodal learning with missing modality: A survey},
  author={Wu, Renjie and Wang, Hu and Chen, Hsiang-Ting and Carneiro, Gustavo},
  journal={arXiv preprint arXiv:2409.07825},
  year={2024}
}

@article{lv2026spd,
  title={SPD-Faith Bench: Diagnosing and Improving Faithfulness in Chain-of-Thought for Multimodal Large Language Models},
  author={Lv, Weijiang and Feng, Yaoxuan and Xia, Xiaobo and Wang, Jiayu and Jing, Yan and Chen, Wenchao and Chen, Bo},
  journal={arXiv preprint arXiv:2602.07833},
  year={2026}
}

@article{gao2026omnimodal,
  title={Omnimodal Dataset Distillation via High-order Proxy Alignment},
  author={Gao, Yuxuan and Liu, Xiaohao and Xia, Xiaobo and Liu, Tongliang},
  journal={arXiv preprint arXiv:2604.10666},
  year={2026}
}

@inproceedings{yun2024flex,
  title={Flex-moe: Modeling arbitrary modality combination via the flexible mixture-of-experts},
  author={Yun, Sukwon and Choi, Inyoung and Peng, Jie and Wu, Yangfan and Bao, Jingxuan and Zhang, Qiyiwen and Xin, Jiayi and Long, Qi and Chen, Tianlong},
  booktitle={NeurIPS},
  pages={98782--98805},
  year={2024}
}

@inproceedings{wang2023incomplete,
  title={Incomplete multimodality-diffused emotion recognition},
  author={Wang, Yuanzhi and Li, Yong and Cui, Zhen},
  booktitle={NeurIPS},
  pages={17117--17128},
  year={2023}
}

@inproceedings{ke2025knowledge,
  title={Knowledge Bridger: Towards Training-Free Missing Modality Completion},
  author={Ke, Guanzhou and He, Shengfeng and Wang, Xiaoli and Wang, Bo and Chao, Guoqing and Zhang, Yuanyang and Xie, Yi and Su, Hexing},
  booktitle={CVPR},
  pages={25864--25873},
  year={2025}
}

@inproceedings{xu2024leveraging,
  title={Leveraging knowledge of modality experts for incomplete multimodal learning},
  author={Xu, Wenxin and Jiang, Hexin and Liang, Xuefeng},
  booktitle={ACM MM},
  pages={438--446},
  year={2024}
}

@inproceedings{zhang2022m3care,
  title={M3care: Learning with missing modalities in multimodal healthcare data},
  author={Zhang, Chaohe and Chu, Xu and Ma, Liantao and Zhu, Yinghao and Wang, Yasha and Wang, Jiangtao and Zhao, Junfeng},
  booktitle={KDD},
  pages={2418--2428},
  year={2022}
}

@article{tang2024modal,
  title={Modal-nexus auto-encoder for multi-modality cellular data integration and imputation},
  author={Tang, Zhenchao and Chen, Guanxing and Chen, Shouzhi and Yao, Jianhua and You, Linlin and Chen, Calvin Yu-Chian},
  journal={Nature Communications},
  volume={15},
  number={1},
  pages={9021},
  year={2024},
  publisher={Nature Publishing Group UK London}
}

@inproceedings{wang2023multi,
  title={Multi-modal learning with missing modality via shared-specific feature modelling},
  author={Wang, Hu and Chen, Yuanhong and Ma, Congbo and Avery, Jodie and Hull, Louise and Carneiro, Gustavo},
  booktitle={CVPR},
  pages={15878--15887},
  year={2023}
}

@inproceedings{chen2024probabilistic,
  title={Probabilistic conformal distillation for enhancing missing modality robustness},
  author={Chen, Mengxi and Zhang, Fei and Zhao, Zihua and Yao, Jiangchao and Zhang, Ya and Wang, Yanfeng},
  booktitle={CVPR},
  pages={36218--36242},
  year={2024}
}

@article{reza2024robust,
  title={Robust multimodal learning with missing modalities via parameter-efficient adaptation},
  author={Reza, Md Kaykobad and Prater-Bennette, Ashley and Asif, M Salman},
  journal={IEEE Transactions on Pattern Analysis and Machine Intelligence},
  year={2024},
  publisher={IEEE}
}

@article{li2025simmlm,
  title={SimMLM: A Simple Framework for Multi-modal Learning with Missing Modality},
  author={Li, Sijie and Chen, Chen and Han, Jungong},
  journal={arXiv preprint arXiv:2507.19264},
  year={2025}
}

@inproceedings{lao2025boosting,
  title={Boosting Discriminability for Robust Multimodal Entity Linking with Visual Modality Missing},
  author={Lao, Mingrui and Li, Zheng and Guo, Yanming and Zhang, Xueyi and Cai, Siqi and Ding, Zhaoyun and Li, Haizhou},
  booktitle={SIGIR},
  pages={989--999},
  year={2025}
}

@inproceedings{wang2023learnable,
  title={Learnable cross-modal knowledge distillation for multi-modal learning with missing modality},
  author={Wang, Hu and Ma, Congbo and Zhang, Jianpeng and Zhang, Yuan and Avery, Jodie and Hull, Louise and Carneiro, Gustavo},
  booktitle={MICCAI},
  pages={216--226},
  year={2023}
}

@inproceedings{ma2021smil,
  title={Smil: Multimodal learning with severely missing modality},
  author={Ma, Mengmeng and Ren, Jian and Zhao, Long and Tulyakov, Sergey and Wu, Cathy and Peng, Xi},
  booktitle={AAAI},
  volume={35},
  number={3},
  pages={2302--2310},
  year={2021}
}

@inproceedings{ma2022multimodal,
  title={Are multimodal transformers robust to missing modality?},
  author={Ma, Mengmeng and Ren, Jian and Zhao, Long and Testuggine, Davide and Peng, Xi},
  booktitle={CVPR},
  pages={18177--18186},
  year={2022}
}

@inproceedings{zhao2021missing,
  title={Missing modality imagination network for emotion recognition with uncertain missing modalities},
  author={Zhao, Jinming and Li, Ruichen and Jin, Qin},
  booktitle={ACL},
  pages={2608--2618},
  year={2021}
}

@inproceedings{jin2023rethinking,
  title={Rethinking missing modality learning from a decoding perspective},
  author={Jin, Tao and Cheng, Xize and Li, Linjun and Lin, Wang and Wang, Ye and Zhao, Zhou},
  booktitle={MM},
  pages={4431--4439},
  year={2023}
}

@inproceedings{radford2021learning,
  title={Learning transferable visual models from natural language supervision},
  author={Radford, Alec and Kim, Jong Wook and Hallacy, Chris and Ramesh, Aditya and Goh, Gabriel and Agarwal, Sandhini and Sastry, Girish and Askell, Amanda and Mishkin, Pamela and Clark, Jack and others},
  booktitle={ICML},
  pages={8748--8763},
  year={2021}
}

@inproceedings{jamal2025multi,
  title={Multi-Modal Contrastive Masked Autoencoders: A Two-Stage Progressive Pre-training Approach for RGBD Datasets},
  author={Jamal, Muhammad Abdullah and Mohareri, Omid},
  booktitle={CVPR},
  pages={17947--17957},
  year={2025}
}

@article{luo2025gui,
  title={Gui-r1: A generalist r1-style vision-language action model for gui agents},
  author={Luo, Run and Wang, Lu and He, Wanwei and Chen, Longze and Li, Jiaming and Xia, Xiaobo},
  journal={arXiv preprint arXiv:2504.10458},
  year={2025}
}

@inproceedings{lei2024vit,
  title={Vit-lens: Towards omni-modal representations},
  author={Lei, Weixian and Ge, Yixiao and Yi, Kun and Zhang, Jianfeng and Gao, Difei and Sun, Dylan and Ge, Yuying and Shan, Ying and Shou, Mike Zheng},
  booktitle={CVPR},
  pages={26647--26657},
  year={2024}
}

@inproceedings{cicchetti2024gramian,
  title={Gramian Multimodal Representation Learning and Alignment},
  author={Cicchetti, Giordano and Grassucci, Eleonora and Sigillo, Luigi and Comminiello, Danilo},
  booktitle={ICLR},
  year={2025}
}

@inproceedings{zhang2021cross,
  title={Cross-modal contrastive learning for text-to-image generation},
  author={Zhang, Han and Koh, Jing Yu and Baldridge, Jason and Lee, Honglak and Yang, Yinfei},
  booktitle={CVPR},
  pages={833--842},
  year={2021}
}

@inproceedings{chen2020simple,
  title={A simple framework for contrastive learning of visual representations},
  author={Chen, Ting and Kornblith, Simon and Norouzi, Mohammad and Hinton, Geoffrey},
  booktitle={ICML},
  pages={1597--1607},
  year={2020}
}

@article{xu2021videoclip,
  title={Videoclip: Contrastive pre-training for zero-shot video-text understanding},
  author={Xu, Hu and Ghosh, Gargi and Huang, Po-Yao and Okhonko, Dmytro and Aghajanyan, Armen and Metze, Florian and Zettlemoyer, Luke and Feichtenhofer, Christoph},
  journal={arXiv preprint arXiv:2109.14084},
  year={2021}
}

@article{luo2022clip4clip,
  title={Clip4clip: An empirical study of clip for end to end video clip retrieval and captioning},
  author={Luo, Huaishao and Ji, Lei and Zhong, Ming and Chen, Yang and Lei, Wen and Duan, Nan and Li, Tianrui},
  journal={Neurocomputing},
  volume={508},
  pages={293--304},
  year={2022},
  publisher={Elsevier}
}

@inproceedings{elizalde2023clap,
  title={Clap: learning audio concepts from natural language supervision},
  author={Elizalde, Benjamin and Deshmukh, Soham and Al Ismail, Mahmoud and Wang, Huaming},
  booktitle={ICASSP},
  pages={1--5},
  year={2023}
}

@inproceedings{zhang2022pointclip,
  title={Pointclip: Point cloud understanding by clip},
  author={Zhang, Renrui and Guo, Ziyu and Zhang, Wei and Li, Kunchang and Miao, Xupeng and Cui, Bin and Qiao, Yu and Gao, Peng and Li, Hongsheng},
  booktitle={CVPR},
  pages={8552--8562},
  year={2022}
}

@inproceedings{guzhov2022audioclip,
  title={Audioclip: Extending clip to image, text and audio},
  author={Guzhov, Andrey and Raue, Federico and Hees, J{\"o}rn and Dengel, Andreas},
  booktitle={ICASSP},
  pages={976--980},
  year={2022}
}

@inproceedings{wu2022wav2clip,
  title={Wav2clip: Learning robust audio representations from clip},
  author={Wu, Ho-Hsiang and Seetharaman, Prem and Kumar, Kundan and Bello, Juan Pablo},
  booktitle={ICASSP},
  pages={4563--4567},
  year={2022}
}

@article{chen2023valor,
  title={Valor: Vision-audio-language omni-perception pretraining model and dataset},
  author={Chen, Sihan and He, Xingjian and Guo, Longteng and Zhu, Xinxin and Wang, Weining and Tang, Jinhui and Liu, Jing},
  journal={arXiv preprint arXiv:2304.08345},
  year={2023}
}

@article{guo2023point,
  title={Point-bind \& point-llm: Aligning point cloud with multi-modality for 3d understanding, generation, and instruction following},
  author={Guo, Ziyu and Zhang, Renrui and Zhu, Xiangyang and Tang, Yiwen and Ma, Xianzheng and Han, Jiaming and Chen, Kexin and Gao, Peng and Li, Xianzhi and Li, Hongsheng and others},
  journal={arXiv preprint arXiv:2309.00615},
  year={2023}
}

@article{wang2024freebind,
  title={Freebind: Free lunch in unified multimodal space via knowledge fusion},
  author={Wang, Zehan and Zhang, Ziang and Cheng, Xize and Huang, Rongjie and Liu, Luping and Ye, Zhenhui and Huang, Haifeng and Zhao, Yang and Jin, Tao and Gao, Peng and others},
  journal={arXiv preprint arXiv:2405.04883},
  year={2024}
}

@article{wang2024omnibind,
  title={Omnibind: Large-scale omni multimodal representation via binding spaces},
  author={Wang, Zehan and Zhang, Ziang and Zhang, Hang and Liu, Luping and Huang, Rongjie and Cheng, Xize and Zhao, Hengshuang and Zhao, Zhou},
  journal={arXiv preprint arXiv:2407.11895},
  year={2024}
}

@inproceedings{girdhar2023imagebind,
  title={Imagebind: One embedding space to bind them all},
  author={Girdhar, Rohit and El-Nouby, Alaaeldin and Liu, Zhuang and Singh, Mannat and Alwala, Kalyan Vasudev and Joulin, Armand and Misra, Ishan},
  booktitle={CVPR},
  pages={15180--15190},
  year={2023}
}

@article{zhu2023languagebind,
  title={Languagebind: Extending video-language pretraining to n-modality by language-based semantic alignment},
  author={Zhu, Bin and Lin, Bin and Ning, Munan and Yan, Yang and Cui, Jiaxi and Wang, HongFa and Pang, Yatian and Jiang, Wenhao and Zhang, Junwu and Li, Zongwei and others},
  journal={arXiv preprint arXiv:2310.01852},
  year={2023}
}

@inproceedings{lyu2024unibind,
  title={Unibind: Llm-augmented unified and balanced representation space to bind them all},
  author={Lyu, Yuanhuiyi and Zheng, Xu and Zhou, Jiazhou and Wang, Lin},
  booktitle={CVPR},
  pages={26752--26762},
  year={2024}
}

@inproceedings{chen2023vast,
  title={Vast: A vision-audio-subtitle-text omni-modality foundation model and dataset},
  author={Chen, Sihan and Li, Handong and Wang, Qunbo and Zhao, Zijia and Sun, Mingzhen and Zhu, Xinxin and Liu, Jing},
  booktitle={NeurIPS},
  pages={72842--72866},
  year={2023}
}

@inproceedings{li2022blip,
  title={Blip: Bootstrapping language-image pre-training for unified vision-language understanding and generation},
  author={Li, Junnan and Li, Dongxu and Xiong, Caiming and Hoi, Steven},
  booktitle={ICML},
  pages={12888--12900},
  year={2022}
}

@inproceedings{bain2021frozen,
  title={Frozen in time: A joint video and image encoder for end-to-end retrieval},
  author={Bain, Max and Nagrani, Arsha and Varol, G{\"u}l and Zisserman, Andrew},
  booktitle={CVPR},
  pages={1728--1738},
  year={2021}
}

@inproceedings{wang2019vatex,
  title={Vatex: A large-scale, high-quality multilingual dataset for video-and-language research},
  author={Wang, Xin and Wu, Jiawei and Chen, Junkun and Li, Lei and Wang, Yuan-Fang and Wang, William Yang},
  booktitle={ICCV},
  pages={4581--4591},
  year={2019}
}

@inproceedings{zhao2024videoprism,
  title={VideoPrism: A Foundational Visual Encoder for Video Understanding},
  author={Zhao, Long and Gundavarapu, Nitesh Bharadwaj and Yuan, Liangzhe and Zhou, Hao and Yan, Shen and Sun, Jennifer J and Friedman, Luke and Qian, Rui and Weyand, Tobias and Zhao, Yue and others},
  booktitle={ICML},
  pages={60785--60811},
  year={2024},
  organization={PMLR}
}

@article{wang2022internvideo,
  title={Internvideo: General video foundation models via generative and discriminative learning},
  author={Wang, Yi and Li, Kunchang and Li, Yizhuo and He, Yinan and Huang, Bingkun and Zhao, Zhiyu and Zhang, Hongjie and Xu, Jilan and Liu, Yi and Wang, Zun and others},
  journal={arXiv preprint arXiv:2212.03191},
  year={2022}
}

@inproceedings{chen2011collecting,
  title={Collecting highly parallel data for paraphrase evaluation},
  author={Chen, David and Dolan, William B},
  booktitle={ACL},
  pages={190--200},
  year={2011}
}

@inproceedings{anne2017localizing,
  title={Localizing moments in video with natural language},
  author={Anne Hendricks, Lisa and Wang, Oliver and Shechtman, Eli and Sivic, Josef and Darrell, Trevor and Russell, Bryan},
  booktitle={ICCV},
  pages={5803--5812},
  year={2017}
}

@inproceedings{krishna2017dense,
  title={Dense-captioning events in videos},
  author={Krishna, Ranjay and Hata, Kenji and Ren, Frederic and Fei-Fei, Li and Carlos Niebles, Juan},
  booktitle={ICCV},
  pages={706--715},
  year={2017}
}

@inproceedings{kim2019audiocaps,
  title={Audiocaps: Generating captions for audios in the wild},
  author={Kim, Chris Dongjoo and Kim, Byeongchang and Lee, Hyunmin and Kim, Gunhee},
  booktitle={ACL},
  pages={119--132},
  year={2019}
}

@inproceedings{drossos2020clotho,
  title={Clotho: An audio captioning dataset},
  author={Drossos, Konstantinos and Lipping, Samuel and Virtanen, Tuomas},
  booktitle={ICASSP},
  pages={736--740},
  year={2020},
  organization={IEEE}
}

@article{sun2023eva,
  title={Eva-clip: Improved training techniques for clip at scale},
  author={Sun, Quan and Fang, Yuxin and Wu, Ledell and Wang, Xinlong and Cao, Yue},
  journal={arXiv preprint arXiv:2303.15389},
  year={2023}
}

@inproceedings{chen2023beats,
  title={BEATs: Audio Pre-Training with Acoustic Tokenizers},
  author={Chen, Sanyuan and Wu, Yu and Wang, Chengyi and Liu, Shujie and Tompkins, Daniel and Chen, Zhuo and Che, Wanxiang and Yu, Xiangzhan and Wei, Furu},
  booktitle={ICML},
  pages={5178--5193},
  year={2023},
  organization={PMLR}
}

@inproceedings{liu2025continual,
  title={Continual multimodal contrastive learning},
  author={Liu, Xiaohao and Xia, Xiaobo and Ng, See-Kiong and Chua, Tat-Seng},
  booktitle={NeurIPS},
  year={2025}
}

@inproceedings{nagrani2022learning,
  title={Learning audio-video modalities from image captions},
  author={Nagrani, Arsha and Seo, Paul Hongsuck and Seybold, Bryan and Hauth, Anja and Manen, Santiago and Sun, Chen and Schmid, Cordelia},
  booktitle={ECCV},
  pages={407--426},
  year={2022}
}

@inproceedings{ngiam2011multimodal,
  title={Multimodal deep learning.},
  author={Ngiam, Jiquan and Khosla, Aditya and Kim, Mingyu and Nam, Juhan and Lee, Honglak and Ng, Andrew Y and others},
  booktitle={ICML},
  volume={11},
  pages={689--696},
  year={2011}
}

@article{xu2023multimodal,
  title={Multimodal learning with transformers: A survey},
  author={Xu, Peng and Zhu, Xiatian and Clifton, David A},
  journal={IEEE Transactions on Pattern Analysis and Machine Intelligence},
  volume={45},
  number={10},
  pages={12113--12132},
  year={2023},
  publisher={IEEE}
}

@article{lu2023theory,
  title={A theory of multimodal learning},
  author={Lu, Zhou},
  journal={NeurIPS},
  volume={36},
  pages={57244--57255},
  year={2023}
}

@article{zong2024self,
  title={Self-supervised multimodal learning: A survey},
  author={Zong, Yongshuo and Mac Aodha, Oisin and Hospedales, Timothy},
  journal={IEEE Transactions on Pattern Analysis and Machine Intelligence},
  year={2024},
  publisher={IEEE}
}

@article{dufumier2024align,
  title={What to align in multimodal contrastive learning?},
  author={Dufumier, Benoit and Castillo-Navarro, Javiera and Tuia, Devis and Thiran, Jean-Philippe},
  journal={ICLR},
  year={2025}
}

@article{liu2026principled,
  title={Principled multimodal representation learning},
  author={Liu, Xiaohao and Xia, Xiaobo and Ng, See-Kiong and Chua, Tat-Seng},
  journal={IEEE Transactions on Pattern Analysis and Machine Intelligence},
  year={2026},
  publisher={IEEE}
}

@article{cao2025generalized,
  title={Generalized domain prompt learning for accessible scientific vision-language models},
  author={Cao, Qinglong and Chen, Yuntian and Lu, Lu and Sun, Hao and Zeng, Zhengzhong and Yang, Xiaokang and Zhang, Dongxiao},
  journal={Nexus},
  volume={2},
  number={2},
  year={2025}
}

@inproceedings{liang2021multibench,
  title={Multibench: Multiscale benchmarks for multimodal representation learning},
  author={Liang, Paul Pu and Lyu, Yiwei and Fan, Xiang and Wu, Zetian and Cheng, Yun and Wu, Jason and Chen, Leslie and Wu, Peter and Lee, Michelle A and Zhu, Yuke and others},
  booktitle={NeurIPS},
  year={2021}
}

@inproceedings{liang2023quantifying,
  title={Quantifying \& modeling multimodal interactions: An information decomposition framework},
  author={Liang, Paul Pu and Cheng, Yun and Fan, Xiang and Ling, Chun Kai and Nie, Suzanne and Chen, Richard and Deng, Zihao and Allen, Nicholas and Auerbach, Randy and Mahmood, Faisal and others},
  booktitle={NeurIPS},
  pages={27351--27393},
  year={2023}
}

@article{xin2025lumina,
  title={Lumina-dimoo: An omni diffusion large language model for multi-modal generation and understanding},
  author={Xin, Yi and Qin, Qi and Luo, Siqi and Zhu, Kaiwen and Yan, Juncheng and Tai, Yan and Lei, Jiayi and Cao, Yuewen and Wang, Keqi and Wang, Yibin and others},
  journal={arXiv preprint arXiv:2510.06308},
  year={2025}
}

@inproceedings{huh2024position,
  title={Position: The platonic representation hypothesis},
  author={Huh, Minyoung and Cheung, Brian and Wang, Tongzhou and Isola, Phillip},
  booktitle={ICML},
  year={2024}
}

@inproceedings{tjandrasuwita2025understanding,
  title={Understanding the emergence of multimodal representation alignment},
  author={Tjandrasuwita, Megan and Ekbote, Chanakya and Ziyin, Liu and Liang, Paul Pu},
  booktitle={ICML},
  year={2025}
}

@inproceedings{zolfaghari2021crossclr,
  title={Crossclr: Cross-modal contrastive learning for multi-modal video representations},
  author={Zolfaghari, Mohammadreza and Zhu, Yi and Gehler, Peter and Brox, Thomas},
  booktitle={ICCV},
  pages={1450--1459},
  year={2021}
}

@inproceedings{giordano2025triangle,
  author       = {Giordano Cicchetti and
                  Eleonora Grassucci and
                  Danilo Comminiello},
  title        = {A {TRIANGLE} Enables Multimodal Alignment Beyond Cosine Similarity},
  booktitle={NeurIPS},
  year         = {2025}
}

@inproceedings{letian2024atouch,
  author       = {Letian Fu and
                  Gaurav Datta and
                  Huang Huang and
                  William Chung{-}Ho Panitch and
                  Jaimyn Drake and
                  Joseph Ortiz and
                  Mustafa Mukadam and
                  Mike Lambeta and
                  Roberto Calandra and
                  Ken Goldberg},
  title        = {A Touch, Vision, and Language Dataset for Multimodal Alignment},
  booktitle    = {ICML},
  publisher    = {OpenReview.net},
  year         = {2024}
}

@inproceedings{xuan2024miradata,
  author       = {Xuan Ju and
                  Yiming Gao and
                  Zhaoyang Zhang and
                  Ziyang Yuan and
                  Xintao Wang and
                  Ailing Zeng and
                  Yu Xiong and
                  Qiang Xu and
                  Ying Shan},
  editor       = {Amir Globersons and
                  Lester Mackey and
                  Danielle Belgrave and
                  Angela Fan and
                  Ulrich Paquet and
                  Jakub M. Tomczak and
                  Cheng Zhang},
  title        = {MiraData: {A} Large-Scale Video Dataset with Long Durations and Structured
                  Captions},
  booktitle    = {NeurIPS},
  year         = {2024},
}

@inproceedings{kepan2025openvid,
  author       = {Kepan Nan and
                  Rui Xie and
                  Penghao Zhou and
                  Tiehan Fan and
                  Zhenheng Yang and
                  Zhijie Chen and
                  Xiang Li and
                  Jian Yang and
                  Ying Tai},
  title        = {OpenVid-1M: {A} Large-Scale High-Quality Dataset for Text-to-video
                  Generation},
  booktitle    = {ICLR},
  publisher    = {OpenReview.net},
  year         = {2025}
}

@inproceedings{jiaming2024onellm,
  author       = {Jiaming Han and
                  Kaixiong Gong and
                  Yiyuan Zhang and
                  Jiaqi Wang and
                  Kaipeng Zhang and
                  Dahua Lin and
                  Yu Qiao and
                  Peng Gao and
                  Xiangyu Yue},
  title        = {OneLLM: One Framework to Align All Modalities with Language},
  booktitle    = {CVPR},
  pages        = {26574--26585},
  publisher    = {{IEEE}},
  year         = {2024},
}

@inproceedings{yi2024internvid,
  author       = {Yi Wang and
                  Yinan He and
                  Yizhuo Li and
                  Kunchang Li and
                  Jiashuo Yu and
                  Xin Ma and
                  Xinhao Li and
                  Guo Chen and
                  Xinyuan Chen and
                  Yaohui Wang and
                  Ping Luo and
                  Ziwei Liu and
                  Yali Wang and
                  Limin Wang and
                  Yu Qiao},
  title        = {InternVid: {A} Large-scale Video-Text Dataset for Multimodal Understanding
                  and Generation},
  booktitle    = {ICLR},
  publisher    = {OpenReview.net},
  year         = {2024}
}

@inproceedings{siddharth2024omnivec,
  author       = {Siddharth Srivastava and
                  Gaurav Sharma},
  title        = {OmniVec: Learning robust representations with cross modal sharing},
  booktitle    = {WACV},
  pages        = {1225--1237},
  publisher    = {{IEEE}},
  year         = {2024}
}

@inproceedings{weixian2024vit_lens,
  author       = {Weixian Lei and
                  Yixiao Ge and
                  Kun Yi and
                  Jianfeng Zhang and
                  Difei Gao and
                  Dylan Sun and
                  Yuying Ge and
                  Ying Shan and
                  Mike Zheng Shou},
  title        = {{VIT-LENS:} Towards Omni-modal Representations},
  booktitle    = {CVPR},
  pages        = {26637--26647},
  publisher    = {{IEEE}},
  year         = {2024}
}

@inproceedings{jia2009imagenet,
  author       = {Jia Deng and
                  Wei Dong and
                  Richard Socher and
                  Li{-}Jia Li and
                  Kai Li and
                  Li Fei{-}Fei},
  title        = {ImageNet: {A} large-scale hierarchical image database},
  booktitle    = {CVPR},
  pages        = {248--255},
  year         = {2009},
}

@inproceedings{chen2020vggsound,
  title={Vggsound: A large-scale audio-visual dataset},
  author={Chen, Honglie and Xie, Weidi and Vedaldi, Andrea and Zisserman, Andrew},
  booktitle={ICASSP},
  pages={721--725},
  year={2020},
  organization={IEEE}
}

@inproceedings{wu2023large,
  title={Large-scale contrastive language-audio pretraining with feature fusion and keyword-to-caption augmentation},
  author={Wu, Yusong and Chen, Ke and Zhang, Tianyu and Hui, Yuchen and Berg-Kirkpatrick, Taylor and Dubnov, Shlomo},
  booktitle={ICASSP},
  pages={1--5},
  year={2023},
  organization={IEEE}
}

@InProceedings{li2025dpu,
    author    = {Li, Shawn and Gong, Huixian and Dong, Hao and Yang, Tiankai and Tu, Zhengzhong and Zhao, Yue},
    title     = {DPU: Dynamic Prototype Updating for Multimodal Out-of-Distribution Detection},
    booktitle = {CVPR},
    month     = {June},
    year      = {2025},
    pages     = {10193-10202}
}

@article{ghojogh2021factor,
  title={Factor analysis, probabilistic principal component analysis, variational inference, and variational autoencoder: Tutorial and survey},
  author={Ghojogh, Benyamin and Ghodsi, Ali and Karray, Fakhri and Crowley, Mark},
  journal={arXiv preprint arXiv:2101.00734},
  year={2021}
}

@article{tipping1999probabilistic,
  title={Probabilistic principal component analysis},
  author={Tipping, Michael E and Bishop, Christopher M},
  journal={Journal of the Royal Statistical Society Series B: Statistical Methodology},
  volume={61},
  number={3},
  pages={611--622},
  year={1999},
  publisher={Oxford University Press}
}

@article{li2006matrix,
  title={Matrix perturbation theory},
  author={Li, Ren-Cang},
  journal={Handbook of linear algebra},
  pages={15--21},
  year={2006},
  publisher={Chapman and Hall/CRC Boca Raton, FL}
}

@article{woodbury1949stability,
  title={The stability of out-input matrices},
  author={Woodbury, Max A},
  journal={Chicago, IL},
  volume={9},
  pages={3--8},
  year={1949}
}

@article{gupta2025better,
  title={Better Together: Leveraging Unpaired Multimodal Data for Stronger Unimodal Models},
  author={Gupta, Sharut and Sundaram, Shobhita and Wang, Chenyu and Jegelka, Stefanie and Isola, Phillip},
  journal={arXiv preprint arXiv:2510.08492},
  year={2025}
}

@article{soomro2012ucf101,
  title={UCF101: A dataset of 101 human actions classes from videos in the wild},
  author={Soomro, Khurram and Zamir, Amir Roshan and Shah, Mubarak},
  journal={arXiv preprint arXiv:1212.0402},
  year={2012}
}

@inproceedings{gemmeke2017audio,
  title={Audio set: An ontology and human-labeled dataset for audio events},
  author={Gemmeke, Jort F and Ellis, Daniel PW and Freedman, Dylan and Jansen, Aren and Lawrence, Wade and Moore, R Channing and Plakal, Manoj and Ritter, Marvin},
  booktitle={ICASSP},
  pages={776--780},
  year={2017},
  organization={IEEE}
}

@inproceedings{piczak2015esc,
  title={ESC: Dataset for environmental sound classification},
  author={Piczak, Karol J},
  booktitle={ACM MM},
  pages={1015--1018},
  year={2015}
}

@article{shen2025carl,
  title={CARL: Critical Action Focused Reinforcement Learning for Multi-Step Agent},
  author={Shen, Leyang and Zhang, Yang and Ling, Chun Kai and Zhao, Xiaoyan and Chua, Tat-Seng},
  journal={arXiv preprint arXiv:2512.04949},
  year={2025}
}

@article{chen2026emergence,
  title={The emergence of abstract thought in large language models beyond any language},
  author={Chen, Yuxin and Zhao, Yiran and Zhang, Yang and Zhang, An and Kawaguchi, Kenji and Joty, Shafiq and Li, Junnan and Chua, Tat-Seng and Shieh, Michael and Zhang, Wenxuan},
  journal={Advances in Neural Information Processing Systems},
  volume={38},
  pages={9902--9933},
  year={2026}
}

@article{shao2024average,
  title={Average user-side counterfactual fairness for collaborative filtering},
  author={Shao, Pengyang and Wu, Le and Zhang, Kun and Lian, Defu and Hong, Richang and Li, Yong and Wang, Meng},
  journal={ACM Transactions on Information Systems},
  volume={42},
  number={5},
  pages={1--26},
  year={2024},
  publisher={ACM New York, NY}
}

@inproceedings{liu2025towards,
  title={Towards modality generalization: A benchmark and prospective analysis},
  author={Liu, Xiaohao and Xia, Xiaobo and Huang, Zhuo and Ng, See-Kiong and Chua, Tat-Seng},
  booktitle={ACM MM},
  pages={12179--12188},
  year={2025}
}

@article{wu2018mvae,
  title={Multimodal generative models for scalable weakly-supervised learning},
  author={Wu, Mike and Goodman, Noah},
  journal={NeurIPS},
  volume={31},
  year={2018}
}

@inproceedings{suttergeneralized,
  title={Generalized Multimodal ELBO},
  author={Sutter, Thomas M and Daunhawer, Imant and Vogt, Julia E},
  booktitle={ICLR},
  year={2021}
}

@article{pcca,
  title={A probabilistic interpretation of canonical correlation analysis},
  author={Bach, Francis R and Jordan, Michael I},
  year={2005},
  publisher={Technical Report 688, Department of Statistics, University of California~…}
}

@article{gfa,
  title={Group factor analysis},
  author={Klami, Arto and Virtanen, Seppo and Lepp{\"a}aho, Eemeli and Kaski, Samuel},
  journal={IEEE transactions on neural networks and learning systems},
  volume={26},
  number={9},
  pages={2136--2147},
  year={2014},
  publisher={IEEE}
}
  \bibliographystyle{icml2026}

  \renewcommand{\cftsecfont}{\normalsize}
  \renewcommand{\cftsubsecfont}{\normalsize}
  \renewcommand{\cftbeforesecskip}{12pt}
  \renewcommand{\cftbeforesubsecskip}{12pt}

  \addtocontents{toc}{\protect
  \setcounter{tocdepth}{2}}

  \newpage
  \appendix
  \onecolumn
  \begin{center}
    \LARGE \textbf{Appendix}
    \vspace{1em}
  \end{center}
  \tableofcontents
  \let
  \addcontentsline\OriginalAddContentsLine
  {\newpage}{
\section{Theoretical Analysis}

In this section, we detail the proof of Theorem~\ref{thm:anchor_shift} (\textit{cf.}, Appendix~\ref{sec:appendix:proof_anchorshift}), the derivation of the closed-form solution of $\widehat{\boldsymbol{\theta}}$ (\textit{cf.}, Appendix~\ref{sec:appendix:resolving_theta}), the anchor shift after calibration (\textit{cf.}, Appendix~\ref{sec:appendix:alleviate_shift}), and the convergence analysis (\textit{cf.}, Appendix~\ref{sec:appendix:convergence_analysis}). 

\subsection{Proof of Theorem~\ref{thm:anchor_shift}}
\label{sec:appendix:proof_anchorshift}
\textit{Recall.}
\textit{
Let $\mathbf{u}_1$ and $\mathbf{u}_1^\Omega$ be the leading left singular vectors of the full multimodal matrix \(\mathbf{Z}\) and its observed submatrix $\mathbf{Z}^\Omega$, respectively. Define $\sigma_1 = \|\mathbf{Z}\|_2$, $\sigma_1^\Omega = \|\mathbf{Z}^\Omega\|_2$, and  
$\eta := \sqrt{\sum_{m \in \bar{\Omega}} \langle \mathbf{u}_1^\Omega, \mathbf{z}^m \rangle^2}$.
Then the anchor shift satisfies  
\begin{equation}
    \sqrt{2\left(1 - \frac{\sigma_1^\Omega + \eta^2}{\sigma_1}\right)} \;\le\; \underbrace{\|\mathbf{u}_1 - \mathbf{u}_1^\Omega\|}_{\|\Delta\|} \;\le\; \frac{\sqrt{2}\,\|\mathbf{Z}^{\bar{\Omega}}\|_2}{\sqrt{\lambda_1} - \sqrt{\lambda_2}}.\nonumber
\end{equation}
}

\begin{proof}
Let multimodal representation is expressed by the concatenation of different unimodal representations: 
\begin{align}
    \mathbf{Z} & = \left[ \mathbf{z}^{m_1}, \mathbf{z}^{m_2}, \dots, \mathbf{z}^{m_k} \right] \in \mathbb{R}^{d \times k},\\
& = \mathbf{U} \boldsymbol{\Sigma} \mathbf{V}^\top,\\
\mathbf{U}& = [\mathbf{u}_1, \dots, \mathbf{u}_d] \in \mathbb{R}^{d \times d},\\
\boldsymbol{\Sigma} & = \mathrm{diag}(\sigma_1, \dots, \sigma_{\min\{d,k\}}),
\end{align}
where $\sigma_1 \geq \sigma_2 \geq \cdots \geq 0$.
We define the virtual anchor as $\mathbf{u}_1$, which is the consensus direction in the spanned by multimodalities.

According to the Davis–Kahan Theorem \cite{li2006matrix}, we have:
\begin{align}
    \sin \angle(\mathbf{u}_1, \mathbf{u}_1^\Omega) \leq \frac{\|\mathbf{Z}^{\bar{\Omega}}\|_2}{\delta},
    \label{eq:davis_kahan}
\end{align}
where $\delta = \sigma_1 - \sigma_2$ represents the spectral gap and $\|\mathbf{Z}^{\bar{\Omega}}\|_2 = \sigma_{\max}(\mathbf{Z}^{\bar{\Omega}})$ denotes the spectral norm of the submatrix associated with the missing modalities. 

\begin{align}
    \|\bm{\Delta}\| &:= \|\mathbf{u}_1 - \mathbf{u}_1^{\Omega}\| = [(\mathbf{u}_1 - \mathbf{u}_1^{\Omega})^\top (\mathbf{u}_1 - \mathbf{u}_1^{\Omega})]^{-
    \frac{1}{2}} \\
    &= [\|\mathbf{u}_1\|^2 +  \|\mathbf{u}_1^{\Omega}\|^2 - 2\mathbf{u}_1^\top\mathbf{u}_1^{\Omega}]^{-\frac{1}{2}} \\
    &= [2 - 2\cos\theta]^{-\frac{1}{2}}\\
    & = [4 \sin^2\left( {\theta}/{2} \right)]^{-\frac{1}{2}} \quad \quad(1 - \cos\theta = 2\sin^2(\theta/2))\\
    & = 2|\sin(\theta/2)| = 2\sin(\theta/2)\quad\quad (\theta\le\pi/2)\\
    & \le \sqrt{2}\sin(\theta)\\
    & \le \frac{\sqrt{2}\|\mathbf{Z}^{\bar{\Omega}}\|_2}{\sigma_1 - \sigma_2} \quad\quad(\text{Eq.~(\ref{eq:davis_kahan}}))\\
    & = \frac{\sqrt{2}\|\mathbf{Z}^{\bar{\Omega}}\|_2}{\sqrt{\lambda_1} - \sqrt{\lambda_2}} \quad\quad (\sigma_1(\mathbf{Z}) = \lambda_1(\mathbf{G}))\\
    & = \frac{\sqrt{2}\|\mathbf{Z}^{\bar{\Omega}}\|_2}{\mathcal{A}} \quad\quad (\mathcal{A}:=\sqrt{\lambda_1} - \sqrt{\lambda_2})\\
    &\le \frac{\sqrt{2|\bar{\Omega}|}}{\mathcal{A}}\quad\quad (\|\mathbf{Z}^{\bar{\Omega}}\|_2\le \sqrt{|\bar{\Omega}|})
\end{align}
This upper bound indicates that a better alignment can lead to better robustness for the case of missing modality.

Now we derive the lower bound as follows:
\begin{align}
        \|\bm{\Delta}\| &:= \|\mathbf{u}_1 - \mathbf{u}_1^{\Omega}\| = [2-2\cos\theta]^{1/2}\\
        &\ge [2(1-\cos^2\theta)]^{1/2}\\
        & = [2(1-|\langle\mathbf{u}_1, \mathbf{u}_1^{\Omega}\rangle|^2)]^{1/2}\\
        & \ge [2-2\frac{\|\mathbf{u}_1^{\Omega\top}\mathbf{Z}\|^2_2}{\sigma_1}]^{1/2}\\
        & = [2-2\frac{\|\mathbf{u}_1^{\Omega\top}\mathbf{Z}^\Omega\|^2_2 + \|\mathbf{u}_1^{\Omega\top}\mathbf{Z}^{\bar{\Omega}}\|^2_2}{\sigma_1}]^{1/2}\\
        & = [2-2\frac{(\sigma_1^\Omega)^2 + \eta^2}{\sigma_1}]^{1/2} \quad\quad(\eta := \sqrt{\sum_{m\in\bar{\Omega}}\langle \mathbf{u}_1^{\Omega},\mathbf{z}^m\rangle^2})
\end{align}
This lower bound reveals that the anchor shift cannot be arbitrarily small. It is fundamentally limited by how much the missing modalities $\bar{\Omega}$ contribute to the leading singular direction. 
\end{proof}

\subsection{Resolving $\widehat{\theta}$}
\label{sec:appendix:resolving_theta}
We consider the joint distribution of $\mathbf{z}$ and $\boldsymbol{\beta}$:
\begin{align}
    \begin{bmatrix}
\boldsymbol{\beta}_n \\
\mathbf{z}_n
\end{bmatrix}
& \sim \mathcal{N} \left(
\begin{bmatrix}
\mathbf{0} \\
\boldsymbol{\mu}
\end{bmatrix},
\begin{bmatrix}
\mathbf{I} & \mathbf{W}^\top \\
\mathbf{W} & \mathbf{W} \mathbf{W}^\top + \boldsymbol{\Sigma}
\end{bmatrix}
\right), \\
\boldsymbol{\mu} & = [\boldsymbol{\mu}^{1\top}, \dots, \boldsymbol{\mu}^{|\Omega|\top}]^\top\\
\mathbf{W} & = [\mathbf{W}^{1\top}, \dots, \mathbf{W}^{|\mathcal{\Omega}|\top}]^\top\\
\boldsymbol{\Sigma} & = \mathrm{diag}((\sigma^1)^2 \mathbf{I}, \dots, (\sigma^{|\Omega|})^2 \mathbf{I})
\end{align}
The condition distribution is $p(\mathbf{z} \mid \mathbf{x}) = \mathcal{N}(\mathbf{z}; \mathbf{m}, \mathbf{V})$.
\begin{align}
    \mathbf{V} & = \boldsymbol{\Sigma}_{\boldsymbol{\beta}} - \boldsymbol{\Sigma}_{\boldsymbol{\beta}\mathbf{z}}\boldsymbol{\Sigma}_{\mathbf{z}}^{-1}\boldsymbol{\Sigma}_{\mathbf{z}\boldsymbol{\beta}}\\
    & = \mathbf{I} - \mathbf{W}^\top(\mathbf{W}\mathbf{W}^\top + \boldsymbol{\Sigma})^{-1}\mathbf{W}\\
    & = \mathbf{I} - \mathbf{W}^\top[
    \boldsymbol{\Sigma}^{-1} - \boldsymbol{\Sigma}^{-1} \mathbf{W} (\mathbf{I}_K + \mathbf{W}^\top \boldsymbol{\Sigma}^{-1} \mathbf{W})^{-1} \mathbf{W}^\top \boldsymbol{\Sigma}^{-1}
    ]\mathbf{W} \quad\quad\quad(\text{Woodbury matrix identity \cite{woodbury1949stability}})\nonumber\\
    & =  \mathbf{I} - \mathbf{A} + \mathbf{A} (\mathbf{I}_K + \mathbf{A})^{-1} \mathbf{A}\quad (\mathbf{A} = \mathbf{W}^\top \boldsymbol{\Sigma}^{-1} \mathbf{W})\\
    & = \left( \mathbf{I} + \mathbf{W}^\top \boldsymbol{\Sigma}^{-1} \mathbf{W} \right)^{-1} \quad(\mathbf{A} (\mathbf{I} + \mathbf{A})^{-1} = \mathbf{I} - (\mathbf{I} + \mathbf{A})^{-1})\\
    & = \Big[\mathbf{I} + \sum_{m\in\Omega} \frac{1}{(\sigma^m)^2}\mathbf{W}^{m\top}\mathbf{W}^m \Big]^{-1}
\end{align}

\begin{align}
    \mathbf{m} &= \mathbb{E}[\mathbf{z}_n \mid \mathbf{x}_n] \\
    & = \mathbf{0} + \mathbf{W}^\top (\mathbf{W} \mathbf{W}^\top + \boldsymbol{\Sigma})^{-1} (\mathbf{z} - \boldsymbol{\mu})\\
    & = (\mathbf{I} + \mathbf{W}^\top \boldsymbol{\Sigma}^{-1} \mathbf{W})^{-1} \mathbf{W}^\top \boldsymbol{\Sigma}^{-1}(\mathbf{z} - \boldsymbol{\mu})\\
    & = \mathbf{V} \mathbf{W}^\top \boldsymbol{\Sigma}^{-1}(\mathbf{z} - \boldsymbol{\mu})\\
    & = \mathbf{V}\sum_{m\in\Omega} \frac{1}{(\sigma^m)^2}\mathbf{W}^{m\top} (\mathbf{z}^m - \boldsymbol{\mu}^m)
\end{align}

Therefore, we have:
\begin{align}
    \left\{
    \begin{aligned}
        \mathbf{V} &= \Big[\mathbf{I} + \sum_{m\in\Omega} \frac{1}{(\sigma^m)^2}\mathbf{W}^{m\top}\mathbf{W}^m \Big]^{-1},\\
        \mathbf{m} &= \mathbf{V}\sum_{m\in\Omega} \frac{1}{(\sigma^m)^2}\mathbf{W}^{m\top} (\mathbf{z}^m - \boldsymbol{\mu}^m),
    \end{aligned}
    \right.
\end{align}

The objective is updated as:
\begin{align}
        \mathbb{E}[\log p(\mathbf{z}, \boldsymbol{\beta} | \widehat{\boldsymbol{\theta}})] & =\mathbb{E}\Big[ \log p(\boldsymbol{\beta}) + \sum_{m\in\Omega} \log p(\mathbf{z}^{m} \mid \boldsymbol{\beta}, \widehat{\boldsymbol{\theta}})\Big]\\
        &= \mathbb{E}\Big[
        -\frac{1}{2}\boldsymbol{\beta}^\top \boldsymbol{\beta} + \sum_{m\in\Omega}\Big[-\frac{d}{2}\log(\sigma^m)^2 - \frac{1}{2(\sigma^m)^2}\mathbb{E}[\|\mathbf{z}^m-\boldsymbol{\mu}^m - \mathbf{W}^m\boldsymbol{\beta}\|^2]\Big] + c
        \Big]
\end{align}

Let us compute the closed-form solution of $\boldsymbol{\mu}^m$ and $\mathbf{W}^m$ according to its related terms. 
\begin{align}
\mathbb{E} \| \mathbf{z}^{m}- \boldsymbol{\mu}^{m} - \mathbf{W}^{m} \boldsymbol{\beta} \|^2 \nonumber& = \| \mathbf{z}^{m} - \boldsymbol{\mu}^{m} - \mathbf{W}^{m} \mathbf{m} \|^2 + \mathrm{Tr}[ \mathbf{W}^{m\top} \mathbf{W}^{m} \mathbf{V}]\\
&\quad \to\ \boldsymbol{\mu}^m = \frac{1}{N}\sum_n (\mathbf{z}^m_n - \mathbf{W}^m\mathbf{m}_n)\\
&\quad \to\ \mathbf{W}^m = \left( \sum_{n=1}^N (\mathbf{z}_n^{m} - \boldsymbol{\mu}^{m}) \mathbf{m}_n^\top \right) \left( \sum_{n=1}^N \mathbb{E}[\boldsymbol{\beta}_n \boldsymbol{\beta}_n^\top] \right)^{-1}
\end{align}

We can also calculate the solution for $\sigma_m^2$:
\begin{align}
    \sigma_m^2 = \frac{1}{N d} \sum_{n=1}^N \left[ \| \mathbf{z}_n^{m} - \boldsymbol{\mu}^{m} - \mathbf{W}^{m} \mathbf{m}_n \|^2+\mathrm{Tr}[ \mathbf{W}^{m\top} \mathbf{W}^{m} \mathbf{V}] \right]
\end{align}

Therefore, for missing modalities $m'\in \mathcal{M}/\Omega$ conditioned on the previous observations, we have:
\begin{align}
    \mathbf{z}^{m'} & = \mathbf{W}^{m'}\bar{\boldsymbol{\beta}} + \boldsymbol{\mu}^{m'} + \bar{\boldsymbol{\epsilon}}^m\\
    & = \mathbf{W}^{m'}\mathbb{E}[\boldsymbol{\beta} | X] + \boldsymbol{\mu}^{m'} + \mathbb{E}[\boldsymbol{\epsilon}^m | X]\\
    & = \mathbf{W}^{m'}\mathbf{m} + \boldsymbol{\mu}^{m'} 
\end{align}

\subsection{Alleviating Anchor Shift}
\label{sec:appendix:alleviate_shift}

Let $\widehat{\mathbf{Z}}^{\bar{\Omega}} = [\mathbf{W}^{m'}\mathbf{m} +\mathbf{\mu}^{m'}|m\in\mathcal{M}/\Omega]$. 
\begin{align}
    \|\Delta_{\text{cal}}\| & = \|\mathbf{u}_1 - \mathbf{u}_1^{\text{cal}}\| \\
    & \le \frac{\sqrt{2}\|[\mathbf{Z}^\Omega,\widehat{\mathbf{Z}}^{\bar{\Omega}}] - \mathbf{Z}\|_2}{\sigma_1 - \sigma_2}\\
    & \le \frac{\sqrt{2}\|[\mathbf{Z}^\Omega,\widehat{\mathbf{Z}}^{\bar{\Omega}}] - \mathbf{Z}\|_F}{\sigma_1 - \sigma_2}\\
    & \le \frac{\sqrt{2|\bar{\Omega}|}\varepsilon}{\sigma_1 - \sigma_2} \quad\quad\quad (\|\hat{\mathbf{z}}^{m'} - \hat{\mathbf{z}}^{m'}\|_2\le \varepsilon, \forall m'\in\bar{\Omega})\nonumber
\end{align}
Here $\varepsilon$ represents the imputation error at the representation level.
From Theorem \ref{thm:anchor_shift}, we have:
\begin{align}
    \|\Delta\|\ge\sqrt{2(1-\frac{\sigma_1^\Omega + \eta^2}{\sigma_1})}.
\end{align}
Therefore, we can derive the condition for $\|\Delta_{\text{cal}}\|\le\|\Delta\|$ as follows:
\begin{align}
    \varepsilon < \frac{\sigma_1 - \sigma_2}{\sqrt{|\bar{\Omega}|}} \cdot \sqrt{1 - \frac{\sigma_1^\Omega + \eta^2}{\sigma_1}}.
\end{align}

This also provides a consistent conclusion to alleviate the anchor shift. The calibration is beneficial, especially in the case of strong alignment among modalities ($\uparrow \sigma_1 - \sigma_2$) and greater original shift ($\downarrow \frac{\sigma_1^\Omega + \eta^2}{\sigma_1}$). 

\subsection{Convergence Analysis.}
\label{sec:appendix:convergence_analysis}

\textit{Recall.
Let $\widehat{\boldsymbol{\theta}}^{(t)}$ be the parameters at iteration $t$ applied to the observed-data log-likelihood $L(\widehat{\boldsymbol{\theta}}) = \sum_n \log p(\mathbf{z}_n^\Omega \mid \widehat{\boldsymbol{\theta}})$ under the generative model in Eq.~(\ref{eq:generative_model}). Given the solution of $q$ and $\widehat{\boldsymbol{\theta}}$, $L(\widehat{\boldsymbol{\theta}}^{(t+1)}) \ge L(\widehat{\boldsymbol{\theta}}^{(t)})$.  
}

\begin{proof}
Let ${L}(\widehat{\boldsymbol{\theta}}^{(t+1)})$ be the log-likelihood for parameters at $t+1$ step. 
\begin{align}
    {L}(\widehat{\boldsymbol{\theta}}^{(t+1)}) & = \sum_{n=1}^{N}\log p (\mathbf{z}_n^\Omega | \widehat{\boldsymbol{\theta}}^{(t+1)}) \\
    & = \sum_{n=1}^{N}\log \int p(\mathbf{z}_n^\Omega, \boldsymbol{\beta}_n | \widehat{\boldsymbol{\theta}}^{(t+1)})d\boldsymbol{\beta}_n\\
    & = \sum_{n=1}^{N}\log \mathbb{E}_{q_n}\Big[\frac{p(\mathbf{z}_n^\Omega,\boldsymbol{\beta}_n|\widehat{\boldsymbol{\theta}}^{(t+1)})}{q_n^{(t+1)}(\boldsymbol{\beta}_n)} \Big] \quad \quad \quad (\mathbf{m}_n,\mathbf{V}_n \gets \widehat{\boldsymbol{\theta}}^{(t)})\nonumber\\
    & \ge \sum_{n=1}^{N} \mathbb{E}_{q_n}\Big[\log\frac{p(\mathbf{z}_n^\Omega,\boldsymbol{\beta}_n|\widehat{\boldsymbol{\theta}}^{(t+1)})}{q_n^{(t+1)}(\boldsymbol{\beta}_n)} \Big]\\
    & \ge \sum_{n=1}^{N} \mathbb{E}_{q_n}\Big[\log\frac{p(\mathbf{z}_n^\Omega,\boldsymbol{\beta}_n|\widehat{\boldsymbol{\theta}}^{(t)})}{q_n^{(t+1)}(\boldsymbol{\beta}_n)} \Big] :=\mathcal{Q} \quad \quad \quad (\widehat{\boldsymbol{\theta}}^{(t+1)} = \arg\max_{\widehat{\boldsymbol{\theta}}} \mathcal{Q}(q^{(t+1)}, \widehat{\boldsymbol{\theta}}))\nonumber\\
    & = L(\widehat{\boldsymbol{\theta}}^{(t)}) - \underbrace{\text{KL}[q_n^{(t+1)}(\boldsymbol{\beta}_n) \| p(\boldsymbol{\beta}_n|\mathbf{z}_n^\Omega, \widehat{\boldsymbol{\theta}})]}_{0}\quad \quad \quad (q(\boldsymbol{\beta}) = p(\boldsymbol{\beta} | \mathbf{z}, \widehat{\boldsymbol{\theta}}))\nonumber\\
    & = L(\widehat{\boldsymbol{\theta}}^{(t)})
\end{align}

Therefore, we have ${L}(\widehat{\boldsymbol{\theta}}^*)\ge\cdots\ge{L}(\widehat{\boldsymbol{\theta}}^{(t+1)})\ge{L}(\widehat{\boldsymbol{\theta}}^{(t)})\ge\cdots\ge L(\widehat{\boldsymbol{\theta}}^{(0)})$, indicating our method can converges to a stationary point of ${L}(\widehat{\boldsymbol{\theta}}^*)$. 
\end{proof}

\section{Related Work}
\label{sec:related_work}

\noindent\textbf{Multimodal representation learning.} To integrate multiple modalities, researchers align them within a unified latent space, where one unimodal representation can be retrieved from another when they correspond to a common instance~\cite{lu2023theory, zong2024self, xu2023multimodal, liu2025continual,jamal2025multi,li2025dpu,shen2025carl,chen2026emergence, luo2025gui,liu2025towards}. For example, CLIP~\cite{radford2021learning} pioneers grounding vision in language space through pairwise contrastive learning~\cite{chen2020simple,zolfaghari2021crossclr, zhang2021cross,shao2024average}, inspiring a series of methods for connecting bimodalities~\cite{guzhov2022audioclip,wu2022wav2clip, elizalde2023clap, zhang2022pointclip, xu2021videoclip,luo2022clip4clip}. Recent work goes beyond pairs, adapting the CLIP paradigm to incorporate more modalities~\cite{guo2023point,wang2024freebind,wang2024omnibind,lyu2024unibind}, exemplified by ImageBind~\cite{girdhar2023imagebind} and LanguageBind~\cite{zhu2023languagebind}. Large-scale data collection~\cite{bain2021frozen,nagrani2022learning, zhao2024videoprism,yi2024internvid,xuan2024miradata,letian2024atouch,kepan2025openvid} and architectural advancements~\cite{li2022blip,chen2023valor,chen2023vast,siddharth2024omnivec,weixian2024vit_lens,jiaming2024onellm}, like VAST \cite{chen2023vast}, have furthered progress, though the learning objective remains constrained to pairwise alignment, \textit{i.e.}, aligning one modality with a predefined anchor. GRAM~\cite{cicchetti2024gramian} proposes to learn multimodal representations simultaneously in a geometric manner by minimizing the volume of a parallelotope spanned by the modality vectors. TRIANGLE~\cite{giordano2025triangle} builds on this by minimizing the designed area for three modalities. PMRL~\cite{liu2026principled} establishes principled foundations and proposes maximizing the largest eigenvalue for anchor-free alignment. However, to support its optimization, all modalities must be present to ensure unbiased estimation, unlike prevailing datasets that typically include only two modalities and thus have missing modalities. To endow this emerging paradigm with greater flexibility, we propose calibrating the incomplete alignment with missing modalities for multimodal representation learning.

\noindent\textbf{Learning with missing modalities.} This setting indicates the incomplete data with respect to modalities, necessitating the model to perform nearly as well as when all modalities are present~\cite{wu2024deep, ma2022multimodal}. The prevailing way is modality imputation, which involves filling in the missing information by composing existing modalities and generating absent modalities~\cite{cai2018deep}. Typically, this focuses on generating high-quality raw data for the missing modalities~\cite{cai2018deep, wang2023incomplete} (\textit{e.g.}, using an auxiliary adversarial loss to generate missing images~\cite{cai2018deep}) or, more commonly, crafting the representations in the latent space based on the observed ones~\cite{zhao2021missing, ma2021smil, zhang2022m3care, wang2023multi, jin2023rethinking, tang2024modal, chen2024probabilistic, yun2024flex} (\textit{e.g.}, indexing from previous multimodal interactions~\cite{zhang2022m3care, yun2024flex}). Some studies also employ advanced models for distillation to yield superior modality representations~\cite{ke2025knowledge, lao2025boosting, wang2023learnable}.
Another research line is to design a specialized fusion module to take advantage of available modalities~\cite{xu2024leveraging, li2025simmlm, reza2024robust}. For instance, SimMLM~\cite{li2025simmlm} introduces a gating network to weigh the contribution of experts corresponding to each modality, and \citet{reza2024robust} designs a modulation function to compensate for the missing modalities. 
However, the above methods are mainly specialized for downstream tasks. Learning better multimodal representations with incomplete modalities is not addressed in the fundamental perspective of multimodal alignment. This motivates this work, especially for the emergence of multimodalities in diverse applications. 

We compare CalMRL against two research lines (\textit{i.e.}, classic pCCA and multimodal VAEs) from a technical perspective to highlight their parallels. \textit{Classical methods:} While CalMRL's generative model shares a linear-Gaussian form, we apply it as a lightweight calibration module in the encoded representation space, coupled with a modern SVD-based alignment objective. Classical pCCA/GFA operate on raw features~\cite{pcca, gfa}, whereas CalMRL delegates nonlinearity to pretrained encoders and uses the linear model solely for cross-modal imputation. This decomposition enables our anchor-shift analysis (Theorem~\ref{thm:anchor_shift}, Corollary~\ref{cor:less_anchor_shift}), which is specific to the alignment-via-Gramian paradigm used in PMRL~\cite{liu2026principled} and GRAM~\cite{cicchetti2024gramian}. \textit{Multimodal VAEs:} MVAE~\cite{wu2018mvae} and MoPoE~\cite{suttergeneralized} use amortized variational inference with nonlinear encoder-decoder pairs end-to-end. CalMRL instead uses exact, closed-form inference, trading decoder expressivity for analytical guarantees and training simplicity. Notably, the PoE structure in our posterior (Eq.~\ref{eq:poterior}) is structurally related to MVAE's Gaussian product-of-experts, but utilizes global (rather than instance-dependent) precision contributions.

\section{Implementation Details}

\subsection{Datasets}
\label{sec:appendix:datasets}
We employ the training dataset \textit{VAST-150K}~\cite{cicchetti2024gramian} at the warm-up stage to avoid a cold start. This dataset is a downsized version of VAST-27M~\cite{chen2023vast} and used for previous works~\cite{cicchetti2024gramian, liu2026principled}.  
VAST-27M involves four modalities, \textit{i.e.}, video, audio, caption, and subtitle for each example. 
We also adopt the training splits of \textit{MSR-VTT} \cite{chen2011collecting} (around 180K, where 20 captions for each video) for vision-text and \textit{AudioCaps} \cite{kim2019audiocaps} (around 45K) for audio-text for the missing modality training.
For benchmarking, we adopt several datasets with their test splits, including vision-text datasets \textit{MSR-VTT} \cite{chen2011collecting}, \textit{DiDeMo} \cite{anne2017localizing}, \textit{ActivityNet} \cite{krishna2017dense}, and \textit{VATEX} \cite{wang2019vatex}, and audio-text datasets \textit{AudioCaps} \cite{kim2019audiocaps} and \textit{Clotho} \cite{drossos2020clotho}. All videos are sampled into 8 frames. For the testing splits, the complete modalities are provided \cite{chen2023vast}. The statistics of these benchmarks are shown in Table \ref{tab:statistics_datasets}.

\begin{table}
\centering
\caption{The statistics of datasets.}
\label{tab:statistics_datasets}
\begin{tabular}{l|ccc|cc} 
\toprule
\multirow{2}{*}{Benchmark} & \multicolumn{3}{c|}{Modalities} & \multirow{2}{*}{\#Train} & \multirow{2}{*}{\#Test}  \\
                           & Video & Audio & Text            &                          &                          \\ 
\midrule
MSR-VTT                    &\ding{51}& -     &\ding{51}& 180,000                  & 1,000                     \\
DiDeMo                     &\ding{51}& -     &\ding{51}& -                        & 1,003                     \\
ActivityNet                &\ding{51}& -     &\ding{51}& -                        & 3,987                     \\
VATEX                      &\ding{51}& -     &\ding{51}& -                        & 1,225                     \\
AudioCaps                  & -     &\ding{51}&\ding{51}& 45,178                   & 704                      \\
Clotho                     & -     &\ding{51}&\ding{51}& -                        & 1,045                     \\
\bottomrule
\end{tabular}
\end{table}

\subsection{Model Architecture}
\label{sec:appendix:model_architecture}
We build our model based on VAST \cite{chen2023vast} to ensure a fair comparison rather than advancing the architecture design. Specifically, we construct the vision encoder with EVAClip-ViT-G~\cite{sun2023eva}, where the vision resolution is set to 224$\times$224 pixels. 
BERT is utilized to implement the text encoder with a maximum caption length limited to 40. For subtitles, the maximum length is extended to 70.
BEATs model~\cite{chen2023beats} is adopted for audio encoding. Each audio is preprocessed into 63 mel-frequency bins, outputting 1024 frames. We also built different multimodal alignment methods with ImageBind~\cite{girdhar2023imagebind} as the backbone. To facilitate efficient fine-tuning, we appended an additional projector after the backbone. Wherein, we implement the VAST baseline using multiple contrastive losses to align the various modalities. To ensure a fair comparison and rigorous validation, we maintain the identical model architectures and hyperparameter settings across VAST, GRAM, TRIANGLE, PMRL, and CalMRL.
\begin{table*}[t]
\centering
\caption{Multimodal retrieval results~(\%) for models trained on full datasets and sole datasets with respect to Recall@5.}
\label{tab:result_5}
\begin{adjustbox}{width=0.85\textwidth,center}
\begin{tabular}{@{}l|cc|cc|cc|cc|cc|cc@{}}
\toprule
 & \multicolumn{2}{c|}{MSR-VTT} & \multicolumn{2}{c|}{DiDeMo} & \multicolumn{2}{c|}{ActivityNet}  & \multicolumn{2}{c|}{VATEX}   & \multicolumn{2}{c|}{AudioCaps}  & \multicolumn{2}{c}{Clotho}   \\  & T$\rightarrow$V    & V$\rightarrow$T   & T$\rightarrow$V     & V$\rightarrow$T  & T$\rightarrow$V    & V$\rightarrow$T  & T$\rightarrow$V    & V$\rightarrow$T  & T$\rightarrow$A    & A$\rightarrow$T   & T$\rightarrow$A    & A$\rightarrow$T    \\ 
\midrule
 VAST $\uparrow$       & 76.8       & 81.8       & 74.6       & 75.6       & 80.2            & 80.8            & 95.7      & 95.8      & 81.5          & 84.7          & 49.0         & 45.1         \\
 GRAM $\uparrow$        & 78.7       & 81.1       & 75.3       & 75.7       & 80.0            & 80.3            & 96.7      & 96.8      & 82.4          & 83.9          & 47.2         & 43.6         \\
 Triangle $\uparrow$    & 78.4       & 79.9       & 73.9       & 74.9       & 79.3            & 79.4            & 94.5      & 95.3      & 80.4          & 84.1          & 42.8         & 45.0         \\
 PMRL $\uparrow$        & 79.6       & 78.8       & 74.9       & 76.5       & 80.9            & 81.3            & 95.6      & 96.2      & 81.7          & 83.5          & 49.0         & 47.6         \\
 CalMRL $\uparrow$      & 83.0       & 80.5       & 75.6       & 76.2       & 80.7            & 81.6            & 96.7      & 96.7      & 80.8          & 82.4          & 47.0         & 46.0         \\
\bottomrule
\toprule
 VAST$\uparrow^\text{AT}$     & {71.8}     & {72.1}     & {65.8}     & {68.4}     & {72.6}          & {74.2}          & {94.4}    & {94.1}    & {81.2}        & {82.5}        & {47.0}       & {46.5}       \\
 GRAM$\uparrow^\text{AT}$    & {67.1}     & {73.0}     & {70.3}     & {70.2}     & {73.2}          & {74.0}          & {66.7}    & {92.8}    & {84.7}        & {85.1}        & {49.0}       & {46.6}       \\
 Triangle$\uparrow^\text{AT}$ & {76.1}     & {74.3}     & {71.6}     & {73.6}     & {76.2}          & {76.4}          & {95.2}    & {95.2}    & {79.3}        & {81.4}        & {41.1}       & {43.5}       \\
 PMRL$\uparrow^\text{AT}$     & {77.1}     & {73.8}     & {70.4}     & {72.3}     & {77.0}          & {76.1}          & {94.9}    & {94.8}    & {84.5}        & {85.2}        & {49.9}       & {48.0}       \\
 CalMRL$\uparrow^\text{AT}$   & {76.9}     & {74.2}     & {71.7}     & {72.8}     & {78.1}          & {76.7}          & {95.3}    & {94.5}    & {82.8}        & {83.1}        & {45.8}       & {46.5}       \\
\bottomrule
\toprule
  VAST$\uparrow^\text{VT}$     & {77.0}     & {82.1}     & {76.8}     & {76.8}     & {82.0}          & {81.8}          & {96.0}    & {96.7}    & {64.3}        & {66.2}        & {32.8}       & {34.0}       \\
 GRAM$\uparrow^\text{VT}$     & {79.9}     & {82.4}     & {76.6}     & {78.1}     & {80.5}          & {80.6}          & {97.1}    & {96.8}    & {66.2}        & {67.9}        & {34.5}       & {34.4}       \\
 Triangle$\uparrow^\text{VT}$ & {78.8}     & {81.2}     & {75.6}     & {75.8}     & {81.0}          & {80.0}          & {95.3}    & {95.6}    & {61.5}        & {64.2}        & {35.2}       & {34.2}       \\
 PMRL$\uparrow^\text{VT}$     & {79.8}     & {81.3}     & {76.8}     & {76.6}     & {82.3}          & {81.3}          & {96.3}    & {96.2}    & {63.1}        & {66.9}        & {38.7}       & {35.6}       \\
CalMRL $\uparrow^\text{VT}$  & {82.5}     & {80.9}     & {76.9}     & {77.9}     & {82.1}          & {81.0}          & {96.5}    & {97.3}    & {64.9}        & {66.6}        & {39.0}       & {36.1}       \\
\bottomrule
\end{tabular}
\end{adjustbox}
\end{table*}

\begin{table*}[t]
\centering
\caption{Multimodal retrieval results~(\%) for models trained on full datasets and sole datasets with respect to Recall@10.}
\label{tab:result_10}
\begin{adjustbox}{width=0.85\textwidth,center}
\begin{tabular}{@{}l|cc|cc|cc|cc|cc|cc@{}}
\toprule
 & \multicolumn{2}{c|}{MSR-VTT} & \multicolumn{2}{c|}{DiDeMo} & \multicolumn{2}{c|}{ActivityNet}  & \multicolumn{2}{c|}{VATEX}   & \multicolumn{2}{c|}{AudioCaps}  & \multicolumn{2}{c}{Clotho}   \\  & T$\rightarrow$V    & V$\rightarrow$T   & T$\rightarrow$V     & V$\rightarrow$T  & T$\rightarrow$V    & V$\rightarrow$T  & T$\rightarrow$V    & V$\rightarrow$T  & T$\rightarrow$A    & A$\rightarrow$T   & T$\rightarrow$A    & A$\rightarrow$T    \\ 
\midrule
 VAST $\uparrow$        & 81.4       & 85.7       & 79.2       & 80.2       & 86.8            & 88.5            & 97.3      & 97.1      & 89.9          & 92.9          & 60.4         & 59.5         \\
 GRAM $\uparrow$        & 86.0       & 86.9       & 80.2       & 81.0       & 86.9            & 87.9            & 98.2      & 98.4      & 90.8          & 94.2          & 60.4         & 57.9         \\
 Triangle $\uparrow$    & 83.8       & 85.1       & 79.1       & 79.6       & 86.8            & 87.4            & 95.8      & 97.1      & 90.1          & 93.6          & 57.0         & 59.1         \\
 PMRL $\uparrow$        & 84.5       & 85.2       & 79.6       & 81.6       & 88.0            & 89.0            & 97.0      & 97.3      & 90.3          & 92.8          & 58.9         & 59.7         \\
 CalMRL $\uparrow$      & 87.3       & 87.1       & 80.5       & 82.0       & 87.9            & 89.5            & 98.1      & 98.0      & 88.2          & 91.3          & 58.5         & 59.1         \\
 \bottomrule
\toprule
VAST$\uparrow^\text{AT}$     & {78.4}     & {79.8}     & {72.3}     & {76.5}     & {79.8}          & {82.5}          & {95.8}    & {95.4}    & {91.2}        & {92.9}        & {60.8}       & {60.0}       \\
 GRAM$\uparrow^\text{AT}$     & {72.4}     & {79.1}     & {74.8}     & {77.2}     & {80.0}          & {82.5}          & {67.1}    & {94.9}    & {93.3}        & {94.3}        & {61.5}       & {60.6}       \\
 Triangle$\uparrow^\text{AT}$ & {82.6}     & {81.1}     & {77.3}     & {79.3}     & {83.1}          & {84.7}          & {96.6}    & {96.3}    & {88.8}        & {90.2}        & {55.1}       & {58.1}       \\
 PMRL$\uparrow^\text{AT}$     & {82.7}     & {80.7}     & {75.9}     & {78.9}     & {84.1}          & {85.2}          & {96.1}    & {96.3}    & {92.0}        & {92.8}        & {61.1}       & {61.8}    \\
 CalMRL$\uparrow^\text{AT}$   & {83.3}     & {81.1}     & {77.2}     & {78.7}     & {85.3}          & {85.7}          & {96.2}    & {96.6}    & {90.5}        & {91.5}        & {59.0}       & {59.0}       \\
\bottomrule
\toprule
VAST$\uparrow^\text{VT}$     & {82.9}     & {86.2}     & {80.9}     & {82.7}     & {88.6}          & {89.2}          & {97.4}    & {97.6}    & {75.9}        & {79.3}        & {43.9}       & {44.6}       \\
 GRAM$\uparrow^\text{VT}$     & {86.9}     & {88.0}     & {80.9}     & {82.9}     & {87.5}          & {88.5}          & {98.4}    & {98.0}    & {76.6}        & {81.4}        & {45.5}       & {43.3}         \\
 Triangle$\uparrow^\text{VT}$ & {84.3}     & {86.8}     & {80.2}     & {80.5}     & {87.9}          & {88.1}          & {97.5}    & {97.5}    & {73.6}        & {78.7}        & {47.0}       & {45.5}       \\
 PMRL$\uparrow^\text{VT}$     & {85.6}     & {87.1}     & {81.4}     & {82.5}     & {88.9}          & {88.8}          & {97.8}    & {97.4}    & {77.1}        & {79.8}        & {49.4}       & {46.8}       \\
 CalMRL $\uparrow^\text{VT}$   & {86.3}     & {87.6}     & {81.9}     & {83.2}     & {89.3}          & {88.9}          & {98.4}    & {98.6}    & {76.6}        & {79.1}        & {49.1}       & {45.8}       \\
\bottomrule
\end{tabular}
\end{adjustbox}
\end{table*}

\subsection{Adaptating Baselines for Missing Modalities}
\label{sec:appendix:adaption_baselines}
For comparison with baselines under the missing-modality scenario, we adapt these methods with the following minor modifications.
For the GRAM \cite{cicchetti2024gramian} and PMRL \cite{liu2026principled} methods, we can still calculate the GRAM matrix for the observed modalities. For instance, if given a V-T dataset, the GRAM matrix $\mathbf{G}$ should be 2-by-2. In this case, GRAM only needs to minimize one singular value, thus approaching PMRL. However, GRAM sets text as the fixed anchor modality by default, which may limit its optimization.
For TRIANGLE \cite{giordano2025triangle}, it is explicitly designed for datasets with three modalities since it optimizes the area of a triangle spanned by the three modalities. Therefore, for datasets with two modalities, we change it to maximize the cosine similarity between the two related modalities for stable training.

\subsection{Warm-up Training}
We employ the warm-up training for initialized generative parameters to avoid the cold-start issue. VAST-150K with complete modalities present is selected. To better mimic the missing modality setting, during the warming-up, we randomly choose one modality as the missing and the others as the observations. We only train the model in one epoch.

\subsection{Hyperparameter Setting}
We adopt the AdamW optimizer for training the parameters with $\beta_1 = 0.9$ and $\beta_2=0.98$. The learning rate is set to $1\times10^{-5}$ and we apply the linear schedule with a warm-up ratio being $0.1$. We set the temperature parameters $\tau=0.05$ and $\tau'=0.1$ and instance matching weight $\alpha$ to $0.1$. All the unimodal representations are transformed with the number of dimensions as 512. The training batch size is set to 64. All the experiments are conducted on the device equipped with 2$\times$NVIDIA H100-80GB GPUs.

\section{Additional Results}
\label{sec:appendix:addtional_results}

\textbf{Comparison on more metrics.} In Tables \ref{tab:result_5} and \ref{tab:result_10}, we report the comprehensive results of multimodal retrieval using the metrics Recall@5 and Recall@10, respectively. A higher top K suggests higher absolute performance and a decreased performance gap. Overall, CalMRL achieves outperforming or comparable results compared to the state-of-the-art methods.

\begin{table}[t]
  \centering
  \caption{Multimodal retrieval results (\%) for different generative methods adapted for the missing modality question.}
  \label{tab:different_generatives}
  \begin{tabular}{lccccccc}
    \toprule
    & {MSRVTT} & {ActivityNet} & {DiDeMo} & {VATEX} & {AudioCaps} & {Clotho} & \textbf{Avg.} \\
    \midrule
    \textbf{CalMRL (Ours)} & 61.1 & 55.4 & 57.1 & 81.3 & 50.1 & 23.8 & \textbf{54.8} \\
    \midrule
    MVAE~\cite{wu2018mvae} & 61.0 & 55.0 & 55.9 & 80.8 & 48.4 & 23.0 & 54.0 \\
    MoPoE~\cite{suttergeneralized} & 61.0 & 55.2 & 55.6 & 80.7 & 48.7 & 23.5 & 54.1 \\
    \midrule
    SMIL~\cite{ma2021smil} & 60.2 & 54.8 & 55.0 & 80.6 & 49.4 & 24.2 & 54.0 \\
    Knowledge Bridger~\cite{ke2025knowledge} & 60.9 & 54.7 & 55.1 & 80.8 & 48.9 & 24.5 & 54.2 \\
    \bottomrule
  \end{tabular}
\end{table}

\textbf{Comparison against diverse methods.} In Table~\ref{tab:different_generatives}, we replace CalMRL's generative model with alternatives (MVAE~\cite{wu2018mvae} and MoPoE~\cite{suttergeneralized}) while keeping everything else fixed. It directly demonstrates that our method is competitive with other alternatives when operating in the same well-structured representation space.
We adapt two representative methods, SMIL~\cite{ma2021smil} and Knowledge Bridger~\cite{ke2025knowledge}, to operate in our representation space and feed into the PMRL alignment pipeline. CalMRL achieves superior results while being simpler and providing theoretical guarantees. We will include these in the revision.

\begin{table}[t]
  \centering
  \caption{Performance comparison on different hyperparameters, including $\tau$, $\tau'$ and $\alpha$}
  \label{tab:hyperparameters}
  \begin{minipage}{0.32\textwidth}
    \centering
    \resizebox{\linewidth}{!}{
    \begin{tabular}{lcccc}
    \toprule
      $\tau$ & \textbf{0.05} & \textbf{0.1} & \textbf{0.2} & \textbf{1} \\
      \midrule
      Avg. R@1 & 54.80 & 54.41 & 54.37 & 54.37 \\
      \bottomrule
    \end{tabular}
    }
  \end{minipage}\hfill
  \begin{minipage}{0.32\textwidth}
    \centering
    \resizebox{\linewidth}{!}{
    \begin{tabular}{lcccc}
    \toprule
      $\tau'$ & \textbf{0.05} & \textbf{0.1} & \textbf{0.2} & \textbf{1} \\
      \midrule
      Avg. R@1 & 54.20 & 54.80 & 54.44 & 54.09 \\
      \bottomrule
    \end{tabular}
    }
  \end{minipage}\hfill
  \begin{minipage}{0.32\textwidth}
    \centering
    \resizebox{\linewidth}{!}{
    \begin{tabular}{lcccc}
    \toprule
      $\alpha$ & \textbf{0} & \textbf{0.1} & \textbf{0.5} & \textbf{1} \\
      \midrule
      Avg. R@1 & 53.30 & 54.80 & 54.41 & 53.21 \\
      \bottomrule
    \end{tabular}
    }
  \end{minipage}
\end{table}

\textbf{Comparison on different hyperparameters.} We report the perfromance of CalMRL \textit{w.r.t.} varing hyperparameters in Table~\ref{tab:hyperparameters}. We observe that performance is relatively stable across a reasonable range of 
$\tau$ and $\tau'$, indicating robustness to their specific settings. 
$\alpha$ shows a clear pattern: removing it entirely or over-weighting it both degrade performance, while moderate values yield comparable results.

\section{Reproducibility}
We provide implementation details, including illustrative algorithm descriptions and flows (see Algorithm \ref{alg:promrl}). The source code will be publicly released for reproducibility.

\section{Limitations}
CalMRL enhances multimodal learning by flexibly handling datasets with missing modalities. Despite its effectiveness and design rationale, this core idea is model-architecture agnostic. Therefore, incorporating a more powerful backbone (such as recent advanced vision-text models) could significantly improve its capabilities.
However, adopting a new backbone necessitates pre-training to adapt these models, which is computationally and data-collection intensive. To this end, we chose a recognized approach, the emerging complete-modality alignment frameworks (\textit{e.g.}, GRAM and PMRL), to validate the effect of CalMRL. }
\end{document}